\documentclass{article}

\usepackage{graphicx} 

\usepackage{hyperref}
\usepackage{amsmath}
\usepackage{amssymb}
\usepackage{multirow}

\usepackage{tikz}
\usetikzlibrary{matrix}

\usepackage[utf8]{inputenc}
\usepackage{scrextend}

\usepackage{graphicx}
\usepackage{subcaption}

\usepackage{colortbl}
\usepackage{pifont}
\usepackage{array}
\usepackage{multirow}
\usepackage{xcolor}

\usepackage{booktabs}

\usepackage{algorithm}
\usepackage{algpseudocode}

\usepackage{mdframed}

\usepackage{empheq}
\usepackage{fancybox}

\usepackage{hyperref}

\usepackage{pgfplots}
\pgfplotsset{compat=1.16} 

\usepackage[title]{appendix}
\usepackage{etoolbox}

\def\email#1{{\texttt{#1}}}
\providecommand{\keywords}[1]
{
  \small	
  \textbf{\textit{Keywords---}} #1
}

\title{Variational Inference via Smoothed Particle Hydrodynamics}
\author{Yongchao Huang \footnote{Author email: \email{yongchao.huang@abdn.ac.uk}}}
\date{May 2024}

\begin{document}

\maketitle

\begin{abstract}
    A new variational inference method, \textit{SPH-ParVI}, based on smoothed particle hydrodynamics (SPH), is proposed for sampling partially known densities (e.g. up to a constant) or sampling using gradients. SPH-ParVI simulates the flow of a fluid under external effects driven by the target density; transient or steady state of the fluid approximates the target density. The continuum fluid is modelled as an interacting particle system (IPS) via SPH, where each particle carries smoothed properties, interacts and evolves as per the Navier-Stokes equations. This mesh-free, Lagrangian simulation method offers fast, flexible, scalable and deterministic sampling and inference for a class of probabilistic models such as those encountered in Bayesian inference and generative modelling. 
\end{abstract}


\section{Introduction}

Particle-based simulations are a type of discretization methods which are used in modelling fluids \cite{miller1989globular,stora1999animating,premoze2003particle,muller2003particle,clavet2005particle,Antuono2010,Shadloo2016}, solids \cite{desbrun1996smoothed,keiser2005unified,bell2005particle} and their interactions \cite{muller2004interaction,han2018sph}. They iteratively track particle motions via Navier-Stokes approximations, and offer approximate, interactive simulations with fast speed, which prevails in computer graphics \cite{Max1981vectorized,Peachey1986modeling}, computer games \cite{gibiansky2019computational} and interactive applications \cite{muller2003particle}. Smoothed-particle hydrodynamics (SPH), a computational method which employs particles to simulate fluid flows, falls under this category \footnote{The other type of discretization methods being mesh-based methods.}. It emerges in recent years as a popular numerical technique for simulating a variety of problems arising in fluid dynamics (e.g. modeling incompressible and compressible flows \cite{sph2019recent}), gas and fire dynamics \cite{Lucy1977numerical,stam1995fire}, astrophysics (e.g. simulating star formation \cite{Gingold1977,mocz2011smoothed} and galaxy collisions \cite{lamb1994galactic}), and engineering (e.g. simulating fluid-structure interactions \cite{han2018sph}, and modeling breaking waves \cite{Roselli201920} and dam breaks \cite{Xu2021}). It divides a fluid into a set of discrete, moving particles; each particle carries physical properties such as mass, position, velocity and density. These properties are smoothed by a kernel function, which approximately restores the continuum fluid. 

SPH is a Lagrangian scheme \footnote{SPH discretizes the Navier-Stokes equations. The smoothing kernel spreads the properties of each particle over a continuum, it is thus an interpolation method for particle system \cite{gibiansky2019computational}. Existence and uniqueness of solution to the Navier-Stokes equations can be found in e.g. \cite{galdi2011introduction}.} for computing smooth field variables such as density, pressure, velocity and forces \cite{sigalotti2021mathematics}. SPH simulation starts by distributing particles across the fluid domain. Each particle is assigned initial properties based on the initial conditions (ICs) of the problem. At a given point, to estimate field quantities, SPH uses properties of nearby particles, weighted by a smoothing kernel which measures their influences. Particles flow under forces (accelerations) induced by, e.g. pressure gradient, viscosity or external forces such as gravity or other body forces, following Newton mechanics. Under certain conditions, e.g. presence of external and internal forces such as viscous friction, they may reach an equilibrium state \cite{mocz2011smoothed}. The computational cost of SPH, as well as the accuracy, primarily depends on the number of particles used \footnote{Other factors impacting the accuracy and computational intensity include the chosen smoothing kernel, the discretization scheme, step size, etc.}.

SPH therefore provides a principled framework for modelling interacting particle systems (IPS) as Newtonian fluids, which motivates its application for probability density sampling and inference: if an external pressure or force filed is applied, particles will move to distribute as per the geometry of this external field. Inspired by this, we apply the SPH principles and devise a particle-based variational inference method, \textit{SPH-ParVI}, which employs a target distribution or its gradient as external pressure/force field and produces particle distribution conforming to the target. These samples, as represented by a transient or converged particle configuration, can then used for tasks such as inferring statistics of the target quantity and making predictions with uncertainties. 

\section{Related work}

\paragraph{SPH} SPH falls into the mesh-free, Lagrangian methods which are developed in late 1970s and 1980s \cite{Max1981vectorized,Peachey1986modeling}.  Gingold and Monaghan \cite{Gingold1977} developed the SPH method, which is applicable to a space of an arbitrary number of dimensions, and applied it to a variety of polytropic stellar models. Monaghan \cite{monaghan1994simulating} extended SPH to deal with free surface incompressible flows. Stam and Fiume first applied SPH to simulate gas and fire phenomena \cite{stam1995fire}. Desbrun and Cani \cite{desbrun1996smoothed} developed an elastic SPH approach for deformable solids. Müller et al. applied SPH to simulate fluids with free surfaces \cite{muller2003particle}, Keiser et al. \cite{keiser2005unified} applied SPH for animating both solids and fluids. Some variants of SPH, e.g. the corrective Smoothed-Particle Method (CSPM) \cite{chen1999improvement,chen1999completeness} and modified CSPM method \cite{zhang2004modified}, have been proposed, particularly to address the issue of particle inconsistency \cite{liu2006restoring,zhou2008accuracy,diBlasi2009consistency,litvinov2015towards,sibilla2015algorithm,zhu2015numerical} which impacts SPH accuracy. A kernel gradient free (KGF) SPH method was proposed by Huang et al. \cite{huang2015kernel,HUANG2019571}. A comprehensive introduction of the SPH method can be found in \cite{Monaghan2005}; insightful summaries of its applications can be found in e.g. \cite{becker2007weakly,Damien2012,Shadloo2016,sph2019recent,nasreldeen2017sph}.

\paragraph{Sampling methods} Sampling from a prescribed distribution, either fully or partially known, is not easy \cite{Bayesian_signal_processing_Joseph}. If we know the full pdf, many statistical sampling methods can be used to generate (quasi) samples. For example, inverse transform sampling \cite{Devroye1986}, rejection sampling \cite{vonNeumann1951}, importance sampling \cite{Kahn1949}, Quasi-Monte Carlo methods (QMC) \cite{Niederreiter1992}, etc. Approximate inference methods provide a pathway to produce (quasi) samples, these include Markov chain Monte Carlo (MCMC) and variational inference (VI) methodologies. MCMC methods such as Metropolis–Hastings (MH) sampling \cite{Metropolis1953}, Gibbs sampling \cite{Geman1984}, Langevin Monte Carlo (LMC) \cite{Roberts1996}, Hamiltonian Monte Carlo (HMC) \cite{Duane1987}, slice sampling \cite{Neal2003}, etc, are commonly used in sampling complex geometries. VI methods such as mean-field VI \cite{Jordan1999introduction}, stochastic VI \cite{Hoffman2013}, black-box VI \cite{Ranganath2014}, variational autoencoders (VAEs) \cite{kingma2013VAE}, etc, are widely used in many applications \cite{Jordan1999introduction,Roberts2002AR}. Particle-based VI (ParVI) methods such as Stein variational gradient descent (SVGD) \cite{Liu2016SVGD}, Sequential Monte Carlo (SMC) \cite{Doucet2001}, particle-based energetic VI (EVI) \cite{Wang2021EVI}, etc, gain popularity due to their efficiency and less suffering from curse of dimensionality. Some of these sampling algorithms utilise the original pdf or up to a constant (i.e. \textit{gradient-free}), e.g. slice sampling and the MH method, while others, e.g. HMC, LMC and most VI methods, utilise the gradient information to guide sampling (i.e. \textit{gradient-based}). Using only gradients enables sampling and inference from an un-normalised, intractable densities which are frequently encountered in Bayesian inference; however, it also increases computational burden. 

Conventional particle-based variational inference (ParVI) methods, e.g. SVGD \cite{Liu2016SVGD}, SMC \cite{Doucet2001}, EVI \cite{Wang2021EVI}, etc, evolve a set of interactive particles towards forming the target geometry by explicitly or implicitly minimizing certain discrepancy measure (e.g. KL divergence), based on statistical principles (e.g. the Stein's identity). Purely physical simulation based approaches, e.g. electrostatics-based ParVI (EParVI \cite{huang2024EParVI}), are recently proposed for variational inference. SPH, in particular, has not been applied in probabilistic machine learning such as Bayesian inference. In the following, we describe the fundamental principles of SPH and demonstrate how it can be applied to statistical sampling and inference; experiments to validate the proposed methodology will be presented in future work.

\section{SPH-based sampling: methodology}

\begin{figure}
    \centering
    \begin{subfigure}{0.3\textwidth}
        \centering
        \includegraphics[width=\linewidth]{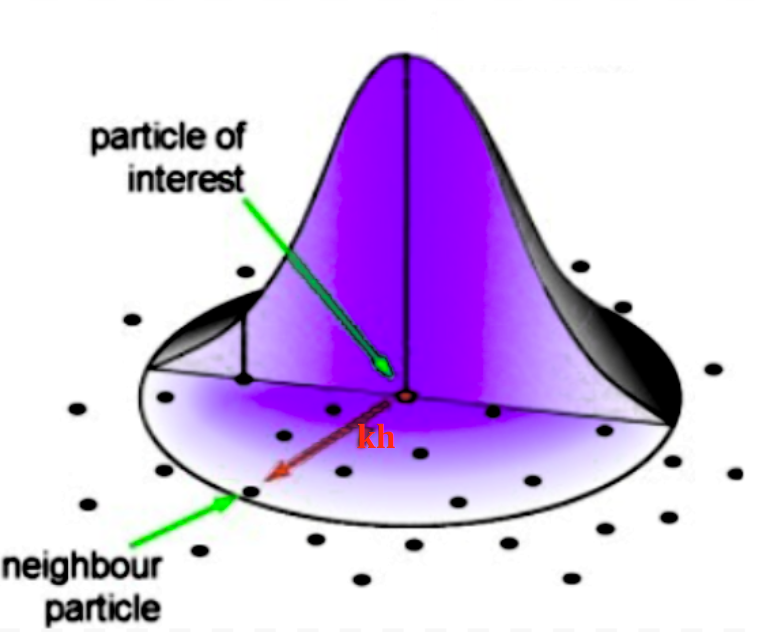}
    \end{subfigure}
    \hspace{0.25cm}
    \begin{subfigure}{0.26\textwidth}
        \centering
        \includegraphics[width=\linewidth]{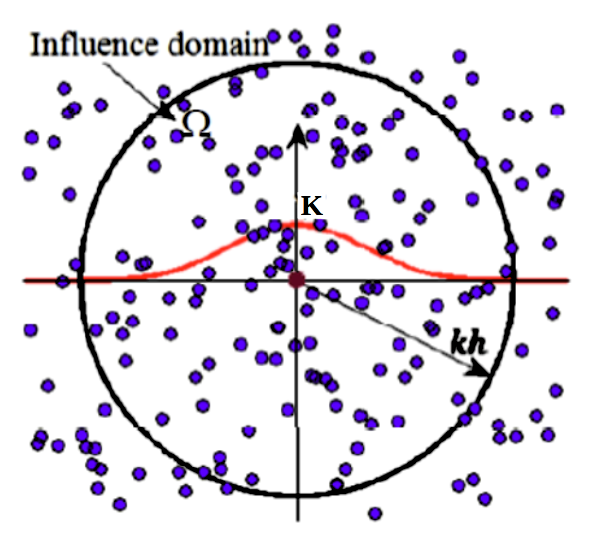}
    \end{subfigure}
    \caption{Illustration of the smoothing kernel (figures from \cite{nasreldeen2017sph}).}
    \label{fig:smoothing_kernel}
\end{figure}

\paragraph{The SPH fluid model} In SPH, particles interpolate a continuous function $f(\textbf{r})$, where $\textbf{r}$ is the position, via a smoothing kernel: 

\begin{equation} \label{eq:cc_integral_discrete_approx} \tag{cc.Eq.\ref{eq:integral_discrete_approx}}
    f(\textbf{r}) = \sum_{j=1}^M \frac{m_j}{\rho_j} f(\textbf{r}_j) K(|\textbf{r}-\textbf{r}_j|, h)
\end{equation}

\noindent where $M$ is the total number of particles used, $m_j$ and $\rho_j$ are the mass and density of the particle $j$, respectively. The smoothing kernel \footnote{In CFD literature, the smoothing kernel is denoted by $W$; here we use $K$ to be consistent with \textit{covariance function} used in statistics literature. Note in general kernels used in SPH are stationary and isotropic (or homogeneous) - it depends on the relative position of its two inputs, and not on their absolute locations \cite{Genton2001kernel}.} $K(r_{ij},h)$, as shown in Fig.\ref{fig:smoothing_kernel},  accepts the distance $r_{ij}=|\textbf{r}_i - \textbf{r}_j|$ as input and computes the weight of influence of the contributing particle $j$ on particle $i$, $h$ is a smoothing length parameter which roughly defines the radius of influence, see Appendix.\ref{app:smoothing_kernel} for some examples. The function $f(\textbf{r})$ being smoothed can be physical quantities such as density, pressure, force, etc. The SPH formulation in \ref{eq:cc_integral_discrete_approx} will be used to approximate these quantities in the Navier-Stokes equations governing fluid motion.

\paragraph{SPH formulation of the Navier-Stokes equations} 

Here we present the SPH description of the interactive particle system for weakly compressible fluids. More details, as well as the SPH formulation for incompressible fluids, can be found in Appendix.\ref{app:SPH}. The Navier-Stokes equations for weakly compressible fluids can be expressed in Lagrangian form as \cite{becker2007weakly,Lind2020}:

\begin{equation} \label{eq:cc_NS_equation_WCSPH} \tag{cc.Eq.\ref{eq:NS_equation_WCSPH}}
    \begin{aligned}
        \frac{d\rho}{dt} &= -\rho \nabla \cdot \mathbf{v} \\
        \frac{d\mathbf{v}}{dt} &= -\frac{1}{\rho} \nabla P + \mathbf{f}^{vis} + \mathbf{f}^{ext}
    \end{aligned}
\end{equation}

\noindent where $\mathbf{v}$ is the velocity, $\rho$ the density, $P$ the pressure, $\mathbf{f}^{vis}$ the viscous force, and $\mathbf{f}^{ext}$ external body forces (e.g. gravitational acceleration \footnote{Strictly speaking, the RHS items are accelerations $\mathbf{a}$ not forces $\mathbf{f}$, they are connected via $\mathbf{a}=\mathbf{f}/m$ where $m$ is the mass. $\mathbf{a}=\mathbf{f}$ when $m=1.0$. We use force over unit mass in this context to imply the source of action.}). $\nabla$ denotes derivative, $\nabla \cdot$ denotes divergence - both with respect to the coordinates (i.e. spatial derivatives). 

The first equation in \ref{eq:cc_NS_equation_WCSPH} is termed \textit{continuity equation} which represents conversation of mass, the second is the \textit{momentum equation} which describes the motion of a fluid. Intuition behind the continuity equation is the continuous mass flow (a strong, local representation of the conservation law); the intuition for the second being, changes in velocity (i.e. the acceleration) of a fluid element is induced by pressure gradient, internal friction (i.e. the resistance force induced by viscosity), and body forces. It essentially echoes the Newton's second law.

\ref{eq:cc_NS_equation_WCSPH} give the governing equations for a continuum fluid; with many benefits \footnote{Of course one is free to use any discretization scheme, e.g. grid-based finite element or difference methods, to solve these PDEs. SPH, however, features less computational cost, easy model implementation and parallelization. An incomplete list of SPH benefits can be found in Section.\ref{sec:discussions}.}, here we use the SPH method which divides the fluid into particles, and applies the SPH smoothing principle to spatially smooth \footnote{SPH assumes the approximated function is sufficiently smooth between fluid elements (e.g. particles).} the physical quantities such as density $\rho$, pressure $P$, and forces $\mathbf{f}^{vis}$ and $\mathbf{f}^{ext}$ at any transient time:

\begin{equation} \label{eq:cc_discrete_NS_equation_WCSPH} \tag{cc.Eq.\ref{eq:discrete_NS_equation_WCSPH}}
\begin{aligned}
    & \frac{d\rho_i}{dt} = \sum_{j=1}^M m_j (\mathbf{v}_i - \mathbf{v}_j)^T \cdot \nabla K_{ij} + \delta_i \quad \text{(Continuity equation)} \\
    & \frac{d\mathbf{v}_i}{dt} = -\sum_{j=1}^M m_j \left( \frac{P_i}{\rho_i^2} + \frac{P_j}{\rho_j^2} \right) \nabla K_{ij} + \mathbf{f}_i^{vis} + \mathbf{f}^{ext} \quad \text{(Momentum equation)}
    \\
\end{aligned}
\end{equation}

\noindent where  $\delta_i$ is an artificial density diffusion term used to limit spurious density fluctuations, see Eq.\ref{eq:artificial_density_diffusion} in Appendix.\ref{app:SPH}. $\mathbf{f}_i^{vis}$ and $\mathbf{f}^{ext}$ are the internal viscous force and external body forces, respectively. The viscous force $\mathbf{f}_i^{vis}$ can have different formulations, see later Eq.\ref{eq:linear_viscosity_force} and Eq.\ref{eq:SPH_viscosity_force1}. The continuity equation is a scalar-valued equation while the momentum equation can be vector-valued as $\mathbf{v}$, $\nabla K$, and $\mathbf{f}$ can be vectors.

The internal pressure $P_i$, induced by change of density during fluid flowing, can be calculated using the equation of state (EoS):

\begin{equation} \label{eq:cc_EoS_pressure_density} \tag{cc.Eq.\ref{eq:EoS_pressure_density}}
    P_i = \frac{c_0^2 \rho_0}{\gamma} [( \frac{\rho_i}{\rho_0} )^\gamma - 1] \quad \text{(Equation of State)} \\
\end{equation}

\noindent where $c_0$ is a reference speed of sound in the fluid, $\rho_0$ the reference density, and $\gamma$ the fluid polytropic index. One observes that, the internal fluid pressure is proportional to local fluid density variations.

Starting from some initial conditions $(\rho_{i}^0,\mathbf{r}_{i}^0,\mathbf{v}_{i}^0)$, \ref{eq:cc_discrete_NS_equation_WCSPH} can be numerically solved using a leapfrog (or Runge-Kutta) time integrator which alternatively updates the position, velocity and density of a particle $i$ within a small time step \(\Delta t\): 

\begin{equation} \label{eq:cc_leapfrog_integrator} \tag{cc.Eq.\ref{eq:leapfrog_integrator}}
\begin{aligned}
    & \mathbf{v}_i^{(t+1)/2} = \mathbf{v}_i^t + \mathbf{a}_i^t \frac{\Delta t}{2}, \\
    & \mathbf{r}_i^{t+1} = \mathbf{r}_i^t + \mathbf{v}_i^{(t+1)/2} \Delta t, \\
    & \mathbf{v}_i^{t+1} = \mathbf{v}_i^{(t+1)/2} + \mathbf{a}_i^{t+1} \frac{\Delta t}{2}, \\ 
    & \rho_i^{t+1} = \rho_i^t + [\sum_{j=1}^M m_j (\mathbf{v}_i^{t+1} - \mathbf{v}_j^{t+1})^T \cdot \nabla K_{ij} + \delta_i] \Delta t \\
\end{aligned}
\end{equation}

\noindent where the superscript $t$ denotes the time step, $\mathbf{a}_i=d \mathbf{v}_i / dt = -\sum_{j=1}^M m_j \left( \frac{P_i}{\rho_i^2} + \frac{P_j}{\rho_j^2} \right) \nabla K_{ij} + \mathbf{f}_i^{vis} + \mathbf{f}^{ext}$ is the acceleration (the RHS of the momentum equation). Calculation of the density $\rho$ can be simplified as the weighted mass of nearby particles:

\begin{equation} \label{eq:simplified_density_cal}
    \rho_i^{t+1} = \sum_{j=1}^M m_j K(|\mathbf{r}_i^{t+1} - \mathbf{r}_j^{t+1}|, h)
\end{equation}

\noindent which can be erroneous for equally distributed surface particles as they have less neighbours \cite{becker2007weakly}.

\paragraph{SPH for sampling} The above SPH procedure simulates the fluid flow with an interacting particle system (IPS) \footnote{Note that, particles interact through the smoothing kernel to ensure continuum physical quantities; they are not real particles in the sense that they do not collide - they might be overlapped though.}. This idea can be adapted to produce (quasi) samples from a target density: the particles can be conceived as samples at a transient or steady state achieved under external conditions (e.g. applied external pressure or force fields, boundary conditions, etc). To realise this, we can apply a target time-invariant, external pressure (scalar) or force (vector) field in the space, i.e. adding $P^{ext}(\mathbf{r}) \propto 1/p(\mathbf{r})$ or $\mathbf{f}^{ext}(\mathbf{r}) \propto \nabla p(\mathbf{r})$ to the momentum equation \footnote{The intuition behind is intuitive: particles shall cluster more in lower pressure and large force regions, yielding high densities.}, and use SPH dynamics to evolve a set of particles toward a transient or equilibrium distribution which approximates the desired, target distribution $p(\mathbf{r})$. We expect that, the clustering of particles, as measured by their densities, would distribute proportional to the external field applied. Physically, starting from any initial configuration, particles, if dominantly driven by external pressure or forces, shall move towards the high-probability regions of the target distribution by design \footnote{Convergence, and potential uniqueness, of the steady state geometry demands more rigorous proof though.}. As such, we provide the two formulations of the proposed SPH-ParVI method in the following.

\paragraph{Representing target density as external pressure field} The first design is applying an time-invariant, external scalar-valued pressure field $P^{ext}(\mathbf{r}):=1/p(\mathbf{r})$ or $P^{ext}(\mathbf{r}):= - \log p(\mathbf{r})$, where $p(\mathbf{r})$ is our target density (known up to a constant \footnote{This is the case in many Bayesian posterior models, resulting from the intractability of the normalising constant.}), to the fluid particle system, and set $\mathbf{f}^{ext}=0$ (or one can conveniently leave $\mathbf{f}^{ext}$=$\textbf{g}$). We therefore re-write the momentum equation as:

\begin{equation} \label{eq:external_pressure_filed_momentum_equation}
    \frac{d\mathbf{v}_i}{dt} = -\sum_{j=1}^M m_j \left( \frac{P_i}{\rho_i^2} + \frac{P_j}{\rho_j^2} \right) \nabla K_{ij} + \mathbf{f}_i^{vis}
\end{equation}

\noindent where the overall point pressure (scalar-valued, as a function of coordinates) acting at particle $i$ becomes $P_i = P_i^{int} + P_i^{ext}$, which is the sum of the internal point pressure $P_i^{int}=\frac{c_0^2 \rho_0}{\gamma} [( \frac{\rho_i}{\rho_0} )^\gamma - 1]$ and external point pressure $P_i^{ext}=1/p(\mathbf{r}_i)$ or $P^{ext}_i= - \log p(\mathbf{r}_i)$.

We are thus simulating a viscous fluid flowing in an unbounded space with free boundary condition, driven dominantly by an external pressure field equaling the the reciprocal of the target distribution (possibly partially known). No other external force presents. Equating the pressure to the inverse (or negative) density is based on the intuition that, small pressure regions shall attract more particles. This design is particularly useful for sampling from a partially known density such as intractable Bayesian posterior. Note that, the internal pressure, resulted from \ref{eq:cc_EoS_pressure_density}, serves as a repulsive force which is necessary when modelling IPS (e.g. for maintaining particle diversity and avoiding mode collapse); it also results in higher pressure in higher particle density regions, and yields bigger repulsive force, which may be undesirable for inferring the target. 

\paragraph{Representing gradient of target density as external force field} The second design is to apply an time-invariant, external vector-valued body force field $\mathbf{f}_i^{ext}:= \nabla_{\mathbf{r}} p(\mathbf{r}) |_{\mathbf{r}=\mathbf{r}_i}$ or $\mathbf{f}_i^{ext}:= \nabla_{\mathbf{r}} \log p(\mathbf{r}) |_{\mathbf{r}=\mathbf{r}_i}$ acting on each particle \footnote{The vector-valued $\nabla_{\mathbf{r}} \log p(\mathbf{r})$ is termed \textit{score field}.}, which revises the momentum equation as:

\begin{equation} \label{eq:external_force_filed_momentum_equation}
\frac{d\mathbf{v}_i}{dt} = -\sum_{j=1}^M m_j \left( \frac{P_i}{\rho_i^2} + \frac{P_j}{\rho_j^2} \right) \nabla K_{ij} + \mathbf{f}_i^{vis} + \nabla_{\mathbf{r}} \log p(\mathbf{\mathbf{r}_i})
\end{equation}

\noindent where the internal pressure $P_i$ remains the same as in \ref{eq:cc_EoS_pressure_density}. We are thus simulating a viscous fluid subject to external body forces equaling the gradients of the target density. This formulation is useful when the gradient information of $p(\mathbf{r})$ or $\log p(\mathbf{r})$ are known, e.g. learned via \textit{score-matching} \cite{hyvarinen2005sm}, and is particularly relevant in scenarios such as diffusion modeling \cite{Ho2020DDPM,scoreSDE2020,SBGM_Huang}. 

In both designs, the target density $p(\mathbf{r})$ doesn't need to be exact; in fact, in most VI problems, $p(\mathbf{r})$ is partially known (up to a constant or only know gradients). One can put a scaling multiplier $\alpha$ attached to the target density or score to proportionally amplify its magnitudes; the magnitudes of the external pressure or force field have impact on the clustering of the particles - large magnitudes can accelerate particle movement, induce oscillations and possibly overlapped particles, while small magnitudes lead to slow convergence. Note that, in both cases, to complete the physics, one can conveniently add a gravitational acceleration $\textbf{g}$ to the RHS of the momentum equation.

As we have free boundary condition, and there is no inflow or outflow, the particle system shall finally achieve equilibrium due to internal friction (i.e. the viscous force $\mathbf{f}^{vis}$), and the stationary distribution of particles, as well certain transient distributions, approaches the target distribution \footnote{Although physically intuitive, this asymptotic behaviour and the invariant distribution statement need some rigor of mathematical proof.}. The SPH sampling procedure, using the external pressure as an example, is described in Algorithm.\ref{algo:SPH_sampling_external_pressure_field}; the case of external force is devised in Algorithm.\ref{algo:SPH_sampling_external_force_field} in Appendix.\ref{app:SPH_sampling_external_force_field}.

\begin{algorithm}[H]
\fontsize{8}{8}
\caption{SPH-based sampling (target density as external pressure field)}
\label{algo:SPH_sampling_external_pressure_field}
\begin{itemize}
\item{\textbf{Inputs}: a queryable target density $p(\textbf{r})$ with magnitude amplification constant $\alpha$, number of particles $M$, particle (density) dimension $d$, initial proposal distribution $p^0(\textbf{r})$, total number of iterations $T$, step size $\Delta t$, smoothing kernel $K(r,h)$ with length-scale $h$. Particle mass $m_i, i=1,2,...,M$. Reference density $\rho_0$, reference speed of sound in the fluid $c_0$, fluid polytropic index $\gamma$, dynamic viscosity $\mu$. Density diffusion parameter $a_d$.} 
\item{\textbf{Outputs}: Particles located at positions $\{\mathbf{r}_i \in \mathbb{R}^d \}_{i=1}^{M}$ whose empirical distribution approximates the target density $p(\textbf{r})$.}
\end{itemize}
\vskip 0.06in
1. \textbf{\textit{Initialise}} particles [$\mathcal{O}(M)$].
    \begin{addmargin}[1em]{0em}%
    Draw $M$ $m$-dimensional particles $\textbf{r}^0=\{\mathbf{r}_j^0\}_{j=1}^{M}$ from the initial proposal distribution $p^0(\textbf{r})$. Initialise $(\rho_{i}^0,\mathbf{r}_{i}^0,\mathbf{v}_{i}^0)$ for i=1,2,...,M. \\ 
    \end{addmargin}

2. \textbf{\textit{Update}} particle positions. \\
For each iteration $t=1,2,...,T$, repeat until converge:
    \\
    \begin{addmargin}[1em]{0em}%
    (1) Compute the density $\rho_i^t$ for each particle $i$=1,2,...,M (Eq.\ref{eq:simplified_density_cal} or \ref{eq:cc_leapfrog_integrator}) [$\mathcal{O}(M^2)$]:
    \[
    \rho_i^t = \sum_{j=1}^M m_j K(|\mathbf{r}_i^t - \mathbf{r}_j^t|, h), 
    \text{ or }
    \rho_i^{t} = \rho_i^{t-1} + [\sum_{j=1}^M m_j (\mathbf{v}_i^{t} - \mathbf{v}_j^{t})^T \cdot \nabla K_{ij} + \delta_i^{t}] \Delta t
    \]
    where $\delta_i^{t}=a_d h c_0 \sum_{j=1}^{M} m_j \frac{{\psi_{ij}^{t}}^T}{\rho_j^{t}} \cdot \nabla K_{ij}$ and $\psi_{ij}^{t} = 2 \left( \frac{\rho_j^{t}}{\rho_i^{t}} - 1 \right) \frac{\mathbf{r}_i^{t} - \mathbf{r}_j^{t}}{|\mathbf{r}_i^{t} - \mathbf{r}_j^{t}|^2 + 0.1 h^2}$. \\ 
    \\
    (2) Query $p(\mathbf{r}_i^t)$ and compute pressure $P_i^t$ for each particle $i$ [$\mathcal{O}(M)$]:
    \[
    P_i^t = \frac{c_0^2 \rho_0}{\gamma} [( \frac{\rho_i^t}{\rho_0} )^\gamma - 1] + 1/[\alpha p(\mathbf{r}_i^t)], \text{ or } P_i^t = \frac{c_0^2 \rho_0}{\gamma} [( \frac{\rho_i^t}{\rho_0} )^\gamma - 1] -\alpha \log [p(\mathbf{r}_i^t) + \epsilon]
    \]
    \noindent where $\epsilon$ is a small value for preventing division by zero. \\ 
    \\
    (3) Compute the overall force $\mathbf{F}_i^t$ acting on each particle $i$ (Eq.\ref{eq:external_pressure_filed_momentum_equation}) [$\mathcal{O}(M^2)$]:  
    \[ 
    \mathbf{F}_i^t = -\sum_{j=1}^M m_j \left( \frac{P_i^t}{\rho_i^{t2}} + \frac{P_j^t}{\rho_j^{t2}} \right) \nabla K_{ij} + \mathbf{f}_i^{vis,t} 
    \]
    where $\mathbf{f}_i^{vis,t}$ is calculated using Eq.\ref{eq:SPH_viscosity_force1} or Eq.\ref{eq:linear_viscosity_force}. \\ 
    \\
    (4) Update $(\mathbf{r}_{i}^t,\mathbf{v}_{i}^t)$ (\ref{eq:cc_leapfrog_integrator} simplified) [$\mathcal{O}(M)$]: \\
    \[
     \mathbf{v}_i^{t+1} = \mathbf{v}_i^t + \frac{\mathbf{F}_i^t}{m_i} \Delta t, \hspace{0.25cm} \mathbf{r}_i^{t+1} = \mathbf{r}_i^t + \mathbf{v}_i^{t+1} \Delta t
    \]    
    \end{addmargin}

3. \textbf{\textit{Return}} the final configuration of particles $\{\mathbf{r}_j^{T}\}_{j=1}^{M}$ and their histogram and/or \textit{KDE} estimate for each dimension. \\
\end{algorithm}

\paragraph{Practical implementation} Based on Algo.\ref{algo:SPH_sampling_external_pressure_field}, in order to optimise efficiency, at each iteration $t$, following procedure can be implemented:

\begin{itemize}
    \item \textit{Step 1}: compute and cache distance matrix $R^t_{M \times M}$ (symmetric, with entries $r^t_{ij}=|\textbf{r}_i^t - \textbf{r}_j^t|$), kernel value matrix $K^t_{M \times M}$ (symmetric, with entries $k^t_{ij}=K(|\textbf{r}_i^t- \textbf{r}_j^t|,h)$), kernel gradient tensor $\nabla K^t_{M \times M \times d}$ (asymmetric, with entries $\nabla k^t_{ij}= \nabla K(|\textbf{r}_i^t - \textbf{r}_j^t|,h)$).
    
    \item \textit{Step 2}: compute particle densities $\{\rho (\textbf{r}_i^t)=\sum_{j=1}^M m_j K(|\mathbf{r}_i^t - \mathbf{r}_j^t|, h)\}_{i=1}^M$ for each particle, and query target value $\{\nabla_{\textbf{r}} \log p(\textbf{r}_i^t) \}_{i=1}^M$ at each particle position.

    \item \textit{Step 3}: retrieve densities and target values, compute for each particle the overall pressure $\{ P_i^t = P_i^{int,t} + P_i^{ext,t} \}_{i=1}^M$, where $P_i^{int,t} = \frac{c_0^2 \rho_0}{\gamma} [( \frac{\rho_i^t}{\rho_0} )^\gamma - 1]$ and $P_i^{ext,t}=-\alpha \log [p(\textbf{r}_i^t) + \epsilon]$.

    \item \textit{Step 4}: retrieve densities, pressures and distances, compute overall force acting on each particle $\{ \textbf{F}_i^t = \textbf{f}_i^{pre,t} +\textbf{f}_i^{vis,t}\}_{i=1}^M$, where $\textbf{f}_i^{pre,t}=-\sum_{j=1}^M m_j \left( \frac{P_i^t}{\rho_i^{t2}} + \frac{P_j^t}{\rho_j^{t2}} \right) \nabla K_{ij}$, $\mathbf{f}_i^{vis,t} = \sum_{j=1}^M \frac{\mu}{\rho_j^t} \frac{m_j}{\rho_j^t} \mathbf{v}_j^t \nabla^2 K(|\textbf{r}_i^t - \textbf{r}_j^t|,h)$ or $\textbf{f}_i^{vis, t}=m_j \eta_{ij} \frac{(\mathbf{v}_i^t - \mathbf{v}_j^t)^T \cdot (\mathbf{r}_i^t - \mathbf{r}_j^t)}{|\mathbf{r}_i^t - \mathbf{r}_j^t|^2 + \epsilon \cdot h^2} \nabla K(|\mathbf{r}_i^t - \mathbf{r}_j^t|, h)$.

    \item \textit{Step 5}: retrieve forces, update velocities and positions $\{ \mathbf{v}_i^{t+1} = \mathbf{v}_i^t + \frac{\mathbf{F}_i^t}{m_i} \Delta t, \hspace{0.25cm} \mathbf{r}_i^{t+1} = \mathbf{r}_i^t + \mathbf{v}_i^{t+1} \Delta t \}_{i=1}^M$.
\end{itemize}

\section{Practical considerations} 

\paragraph{Viscous force} Viscosity is a measure of the fluid's resistance to deformation or flow. In simpler terms, it quantifies how 'sticky' a fluid is. Higher viscosity means the fluid flows less easily. Viscosity is essential for simulating realistic fluid behavior, such as smoothing out velocity gradients and preventing excessive turbulence. Viscous force represents fluid's internal friction which results in resistance of relative motion between particles and slows down particles (so that the simulation can evolve to a steady state). Artificial viscosity can also be introduced to improve the numerical stability and to allow for shock phenomena \cite{Monaghan2005,becker2007weakly}.

When a solid or fluid is subject to shear force, the shear stress $\tau$ is proportional to the spatial gradient of velocity \footnote{In this context, we focus on \textit{Newtonian fluid} whose shear stress is proportional to the shear strain rate. Other fluids which exhibit non-linear relation between shear stress and strain rate include psedoplastic and plastic fluids.}:

\[
\tau \propto \frac{d\mathbf{v}}{d\mathbf{r}}
\]

\noindent the ratio is the \textit{dynamic viscosity} or \textit{absolute viscosity} \(\mu\), i.e. $\tau = \mu \frac{d\mathbf{v}}{d\mathbf{r}}$. $\mu$ negligibly depends on pressure, but may have  different formulations - for liquids it decreases with temperature, for gases it increases with temperature. For fluids, the \textit{kinematic viscosity} or \textit{momentum diffusivity} is often used:

\[
\nu = \frac{\mu}{\rho}
\]

\noindent Kinematic viscosity is a measure of a fluid's resistance to flow and deformation due to internal friction. Again, for gases it increases with temperature, for liquids it decreases with temperature. The viscous force $\textbf{f}^{vis}$ can be calculated as:

\begin{equation} \label{eq:viscosity_force}
    \textbf{f}^{vis} = \nu \nabla^2 \mathbf{v} + \frac{1}{3} \nu \nabla (\nabla \cdot \mathbf{v}) = \frac{\mu}{\rho} \nabla^2 \mathbf{v} + \frac{1}{3} \frac{\mu}{\rho} \nabla (\nabla \cdot \mathbf{v})
\end{equation}

\noindent where $\nabla^2 \mathbf{v}$ is the Laplacian of the velocity vector. The term \(\mathbf{f}_i^{vis}=\nu \nabla^2 \mathbf{v}\) represents the diffusion of momentum due to viscosity, which smooths out velocity gradients within the fluid. It is essential for modeling the dissipative effects of viscosity in fluid flow simulations, ensuring that the simulation accounts for the internal friction that resists the motion of fluid layers relative to each other. The second term $\nabla \cdot \mathbf{v}$ is the divergence of the velocity field which describes the rate at which the fluid is expanding or contracting. There are also linear form of the viscous force, see e.g. Eq.\ref{eq:linear_viscosity_force}, assuming the viscosity coefficient $\nu$ doesn't change with temperature. Also, unless for strongly compressible fluids in which the bulk viscosity term $\frac{1}{3} \frac{\mu}{\rho} \nabla (\nabla \cdot \mathbf{v})$ is needed, one can safely ignore the last term $\frac{1}{3} \frac{\mu}{\rho} \nabla (\nabla \cdot \mathbf{v})$, which gives $\textbf{f}^{vis} = \nu \nabla^2 \mathbf{v} = \frac{\mu}{\rho} \nabla^2 \mathbf{v}$. Assuming constant $\nu$, and apply the SPH smoothing principle, which gives \cite{gibiansky2019computational}:

\begin{equation} \label{eq:SPH_viscosity_force1}
    \mathbf{f}_i^{vis} = \nu \nabla^2 \mathbf{v}(\textbf{r}_i) = \nu \sum_{j=1}^M \frac{m_j}{\rho_j} \mathbf{v}_j \nabla^2 K(|\textbf{r}_i - \textbf{r}_j|,h)
\end{equation}

\noindent or its symmetric variant: 

\begin{equation} \label{eq:SPH_viscosity_force2} \tag{\ref{eq:SPH_viscosity_force1}b}
    \mathbf{f}_i^{vis} = \nu \sum_{j=1}^M \frac{m_j}{\rho_j} (\mathbf{v}_j - \mathbf{v}_i) \nabla^2 K(|\textbf{r}_i - \textbf{r}_j|,h)
\end{equation}

Alternatively, \cite{Monaghan2005,becker2007weakly} used the following \textit{linear} viscous force:

\begin{equation} \label{eq:linear_viscosity_force}
    \mathbf{f}_i^{vis} = 
    \begin{cases} 
    \sum_j m_j \eta_{ij} \frac{(\mathbf{v}_i - \mathbf{v}_j)^T \cdot (\mathbf{r}_i - \mathbf{r}_j)}{|\mathbf{r}_i - \mathbf{r}_j|^2 + \epsilon \cdot h^2} \nabla K(|\mathbf{r}_i - \mathbf{r}_j|, h) & \text{if } \mathbf{v}_{ij}^T \cdot \mathbf{r}_{ij} < 0 \\
    0 & \text{if } \mathbf{v}_{ij}^T \cdot \mathbf{r}_{ij} \geq 0 
    \end{cases}
\end{equation}

\noindent where $\mathbf{v}_{ij}=\mathbf{v}_{i} - \mathbf{v}_{j}$, with \(\mathbf{v}_{ij} \cdot \mathbf{r}_{ij} > 0 \) being equivalent to \( \nabla \cdot \mathbf{v} > 0 \). The viscous coefficient $\eta_{ij}$ is:

\[
\eta_{ij} = \frac{2 \alpha h c_s}{\rho_i + \rho_j}
\]

\noindent in which the viscosity constant \( \alpha \) is between 0.08 and 0.5 as used in \cite{becker2007weakly}. The term \( \epsilon h^2 \) is introduced to avoid singularities for \( |\mathbf{r}_{ij}| = 0 \), with \( \epsilon = 0.01 \). $c_s$ denotes the speed of sound in the fluid. To be consistent with \cite{Lind2020} and \cite{becker2007weakly}, the author suggests $c_s=c_0$ where $c_0$ is the reference speed of sound in the fluid in Eq.\ref{eq:EoS_pressure_density}.

\paragraph{Surface tension} Surface tension plays
an important role in realistic fluid animations \cite{becker2007weakly}, particularly for large curvature geometries (which mapping to our inference problem refers to complex density shape). The main issue is due to the fact that, molecules in a fluid experience intramolecular attraction forces, and molecules in these regions have too few neighbours. For internal particles, the attraction forces are balanced, resulting no net acceleration; for particles on the edge, it is unbalanced and non-negligible. There are several surface tension models to handle surface tension in SPH \cite{Morris1999surface,becker2007weakly}. For example, one of these, which requires calculating the second derivative of a color field, estimates the surface tension force as \cite{gibiansky2019computational}:

\begin{equation}
    f^{sur} = \sigma \kappa \mathbf{n} = - \sigma \nabla \sum_{j=1}^M \frac{m_j}{\rho_j} \nabla^2 K_{ij} \frac{\mathbf{n}}{|\mathbf{n}|}
\end{equation}

\noindent where $\sigma$ is the surface tension coefficient. $\kappa$ is the curvature of the surface. $\mathbf{n}$ is the normal vector to the surface. 

\paragraph{Regularisation} To prevent mode collapse, i.e. local particles collapsing into a single mode point, and encourage particle diversity, one can add a density-based regularization (repulsive) force to the RHS of the momentum equation:

\begin{equation} \label{eq:regularisation_force}
    \mathbf{f}_{i}^{reg} = \sum_{j=1}^M m_j \frac{\nabla K(|\mathbf{r}_i - \mathbf{r}_j|, h)}{\rho_i + \epsilon}
\end{equation}

\noindent where $\epsilon$ is a small number to prevent division by zero.

\paragraph{Smoothing length} The smoothing length $h$ balances local and global smoothness. The power of SPH lies in its adaptive property, i.e. the resolution of a SPH simulation can be adjusted based on local conditions such as density. We can assign time-variant, particle-bespoke (location adaptive) smoothing lengths, allowing particles to adapt itself based on local conditions: (1) Time-varying annealing scheme. In a preliminary experiments, we find that if a constant $h$ is used, however, the overall particle shape expands and contracts cyclically over time. This oscillating behaviour, which is not computationally efficient, can be alleviated by using a time-varying $h$. (2) Spatially-varying $h$. In a dense region where particles are close to each other, $h$ can be made small, which gives high spatial resolution; in low-density regions where particles are far apart, $h$ can be increased so that particles won't get trapped. This also optimises the computational efforts. 

\paragraph{Step size of time} Choosing an appropriate time step is critical for numerical stability and convergence \cite{becker2007weakly}. The following step size, derived based on the Courant-Friedrichs-Lewy(CFL) condition, is suggested \cite{becker2007weakly}:

\begin{equation}
    \Delta t = \min \left( 0.25 \cdot \frac{h}{|\mathbf{f}^{ext}_{min}|}, 0.4 \cdot \frac{h}{c_s \cdot (1 + 0.6 \alpha)} \right)
\end{equation}

\noindent where $\mathbf{f}^{ext}_{min}$ is the minimum external forces among all particles, $h$ the smoothing length, and \( \alpha \) the viscosity constant (0.08 $\sim$ 0.5 as used in \cite{becker2007weakly}). $c_s$ denotes the speed of sound in the fluid.

\paragraph{Boundary conditions} For the inference problem, we simulate particle movement in a free-boundary space, no boundary conditions (BCs) are considered to constrain particle movements. Realistically, boundaries exert a force on particles colliding with a boundary; appropriate boundary conditions, if desired by the inference problem (e.g. inference over a manifold), can be inserted to ensure that the particles behave correctly at the edges of the simulation domain. These might involve reflecting particles at solid boundaries or allowing them to interact with boundaries representing inflow/outflow conditions \cite{Monaghan2005,becker2007weakly}. BCs, as well as initial conditions (ICs), can be accounted for by e.g. applying a constant or time-varying external pressure or force field.

\paragraph{Optimal state} Due to the oscillating behaviour, the optimal state, which is expected to match the target distribution, doesn't necessarily need to be the steady state. We need some notion of dispersion measure which helps define the optimal configuration, ideally we would like particles to cluster around multiple modes, and the optimal configuration may correspond to the step at which the within-mode distances are small while between-mode distances are large (which is similar to optimal clustering criteria). In experiments, one can track the average density (as a simple measure of particle dispersion), and consider the particle configuration at maximum average density to give optimal approximation.

\paragraph{Computational complexity} As noted in Algo.\ref{algo:SPH_sampling_external_pressure_field} (also Algo.\ref{algo:SPH_sampling_external_force_field}), the computational complexity per iteration is $\mathcal{O}(M^2)$ where $M$ is the number of particles used, and total effort over $T$ steps is $\mathcal{O}(M^2 T)$. The complexity increases linearly with extra force terms added (e.g. regularisation). As pairwise distances are used in calculating the kernel and its gradient values, they can be cached. For example, in each step, we can pre-compute pairwise distances and the gradient of the kernel once, and then reuse these values for smoothing densities, pressures, and forces. These efforts, as described in the practical implementation section following the algorithm, avoid redundant calculations and improve efficiency.

\section{Discussions} \label{sec:discussions}

As accuracy and efficiency of the SPH-ParVI method relies on the SPH technique, here we touch some of the topics related to SPH and beyond. 

\paragraph{Advantages of SPH-ParVI} We identify following advantages of using SPH for sampling: (1) \textbf{Advantages of SPH}. This mesh-free particle method enjoys some nice properties of both continuum and fragmentation. Due to its Lagrangian nature, SPH has many advantages over traditional mesh-dependent Eulerian methods \cite{sigalotti2021mathematics}. For example, mesh-based methods may suffer from mesh distortions which affect the numerical accuracy when simulating large material deformations. Also, advection is performed exactly and therefore material history information can be tracked free of numerical diffusion. Further, the mesh-free characteristic significantly simplifies model implementation and its parallelization. (2) \textbf{Physics-based, flexible and expressive interaction modelling}. SPH-ParVI models an interacting particle system (IPS) based on fluid mechanics that preserve mass, momentum and energy. Different types of interactions can be accounted by flexibly introducing corresponding force/pressure terms. (3) \textbf{Computationally efficient}. SPH works independently of any grid (e.g. finite-difference methods), the Lagrangian formulation is more efficient than mesh-based, Eulerian methods - interactions and derivatives are all evaluated in a coordinate system attached to a moving fluid element (i.e. the particle) \cite{mocz2011smoothed}. Efficiency is also improved via evaluating nearby physical quantities through the smoothing kernel: unlike finite element method which solves systems of equations, SPH computes continuum physics (e.g. density, pressure) from weighted contributions of neighbouring particles. The kernel weighted contrition from each particle saves computational effort by excluding the relatively minor contributions from distant particles. (4) \textbf{Deterministic}. SPH-ParVI employs deterministic physical laws to generate samples, the results are trackable and reproducible. (5) \textbf{Flexible and adaptive}. It is flexible to devise location and time dependent resolution in particle methods. SPH's particle-centric approach ensures that computational resources are concentrated in regions where the fluid is present: it automatically adapts the resolution based on the particle distribution, providing higher resolution in regions with more particles (e.g. regions of higher density). One can also design an adaptive (e.g. annealing) smoothing length $h$ to improve sampling quality. Besides, the two formulations of SPH-ParVI provides the flexibility for sampling from partially known densities or from gradients. (6) \textbf{Handling complex geometries}. Mesh-free ParVI methods, e.g. SVGD \cite{Liu2016SVGD,SBGM_Huang}, can handle complex (e.g. large curvature, multi-modal) geometries; for physics-based ParVI methods such as EParVI \cite{huang2024EParVI} and SPH-ParVI, this can be enhanced by introducing boundary conditions. (7) \textbf{Scalable}. SPH method can easily be extended to deal with a large variety of complex physical models \cite{mocz2011smoothed}. Unlike mesh-based, Eulerian methods (e.g. EParVI \cite{huang2024EParVI}), this Lagrangian method suffers less from curse of dimensionality. The flexibility, computability, speed and expressiveness enable SPH-ParVI scalable to complex geometries and large data sets. 

\paragraph{Disadvantages of SPH-ParVI} We identify following disadvantages of using SPH for sampling: (1) \textbf{Disadvantages of SPH} \footnote{Other SPH concerns include its limited accuracy in multi-dimensional flow modelling due to noise, instabilities across contact discontinuities (e.g. Kelvin-Helmholtz instabilities), high artificial viscosity, etc.}. The vanilla SPH method can suffer from slow numerical convergence due to loss of particle consistency \cite{sigalotti2021mathematics}. Mathematically, consistency refers to how well the discretized equations approximate the exact differential equations. \textit{Particle inconsistency} \footnote{Four sources of particle inconsistency have been identified \cite{Rasio2000particle,zhu2015numerical,sigalotti2016kernel,sigalotti2021mathematics}: (1) Kernel support truncation due to the presence of a physical boundary; (2) irregular distributions of particles; (3) spatially varying smoothing lengths in adaptive calculations; (4) use of small numbers of neighbors per particle in simulations with compactly supported kernels.} \cite{sigalotti2021mathematics} arises due to two types of approximations (discretizations) in standard SPH formulation: kernel approximation and particle approximation. Corrective strategies and improved SPH schemes \cite{li1996moving,liu1997moving,bonet1999variational,liu2003constructing,liu2006restoring,zhang2004modified,ferrand2011consistent,ferrand2013unified,reinhardt2019consistent} to restore particle consistency, e.g. the corrective smoothed particle method (CSPM) \cite{chen1999improvement,chen1999completeness}, have been proposed. Also, setting boundary conditions (BCs) in SPH is more difficult than grid-based methods, as near-boundary particles change over time \cite{Shadloo2016}. Further, inaccuracies can be induced due to e.g. loss of symmetry and constant density \cite{gibiansky2019computational}. (2) \textbf{Tuing hyper-parameters}. Key hyper-parameters used in SPH-ParVI are the smoothing length $h$, the target amplification constant $\alpha$, and the discretization time step $\Delta t$. $h$ balances local and global smoothness, while $\alpha$ and $\Delta t$ can affect convergence speed and accuracy. Both have impact on the evolution behaviour of the IPS: in preliminary experiments, we find that group of particles can exhibit contraction and expansion and oscillating behaviours according to different values of $h$, $\alpha$ and $\Delta t$. Therefore, one may need a smart design (e.g. an adaptive annealing scheme) of the hyper-parameters when inferring complex, e.g. multi-modal, densities to achieve good performance.

\paragraph{Mesh-based \textit{vs} meshless} There are two approaches for specifying a flow field: Lagrangian and Eulerian. Both approaches have been long used in computer graphics \cite{becker2007weakly}. Mesh-based approaches such as mass-spring systems, finite element, finite difference, finite volume methods are most widespread. Mesh-free methods, particularly particle systems, are increasingly popular for modelling elasto-plastic materials and fluids \cite{keiser2005unified}. 

Eulerian specification focuses on specific locations in the space through which the fluid flows as time passes. Eulerian simulations use a fixed mesh and computations are performed on a stationary grid - each cell must be computed regardless of it contains significant fluid properties or not, and the cost is therefore linked to the total number of cells. Eulerian approaches can handle inflow and outflow easily by setting boundary conditions \cite{becker2007weakly}. 

The Lagrangian specification looks at fluid motion by following an individual fluid parcel as it moves through space and time. Lagrangian simulations are in general mesh-free and focus on simulation nodes that may move following the velocity field. Lagrangian approaches are more flexible for time-varying simulations and are easy to implement. They are more efficient than grid-based simulation methods, in terms of memory and computation time \cite{becker2007weakly}. Also, mesh-free methods can handle topological changes (e.g. splashes \cite{obrien1995dynamic}) for fluid-like materials, without the need of re-meshing. However, mesh-free methods require the definition of an implicit or explicit surface, and fluid surfaces are smooth due to surface tension. Also, Lagrangian approaches generally require search for neighbours.

SPH is a particle-based, mesh-free Lagrangian approach. In SPH, particles cluster in regions of high fluid density, and yield high resolution in these regions; while in low density regions, the distribution of particles is more sparse. It allocates more particles to focus on areas of interest (i.e. where the fluid is denser), while the resolution of grid-based methods is uniform across the entire grid. Further, it is straightforward for SPH to calculate density-related metrics, as each particle directly represents a portion of the fluid mass and its density; grid-based approaches, however, may require additional efforts, e.g. interpolation, to estimate the density within a cell, which is less efficient and accurate, especially in regions with low fluid content. 

\paragraph{Compressible \textit{vs} incompressible fluids} Some substances such as gas and soft materials (e.g. elastomers, biological tissues) are compressible or partially compressible. For example, the volume of gas can change under pressure due to the usually large empty space between gas particles; liquids are generally incompressible as their molecular structure is compact. Implausible SPH simulations due to compressibility can be resulted \cite{keiser2005unified,bridson2006fluid,becker2007weakly}. Enforcing incompressibility for particle methods is a challenging problem \cite{becker2007weakly}; it can be achieved, at the cost of solving the Poisson equation, by projection approaches \cite{premoze2003particle} which are based on pressure correction. 

The Navier-Stokes equations for an incompressible fluid (e.g. water) require the density to be constant, which for SPH maintaining a constant density is difficult \cite{gibiansky2019computational}. The non-constant density makes the SPH method less accurate in modelling incompressible fluids. While SPH provides speedy and realistic-looking fluid simulation, the density problem may block it from scenarios which require precision or accuracy \cite{gibiansky2019computational}. This, however, can be compensated by using a strong density restoring force (i.e. pressure) \cite{gibiansky2019computational} or using an adaptive smoothing length $h$ to enforce constant density \cite{becker2007weakly}. Alternatively, weakly compressible formulation with very low density fluctuations, as used in this context, has been developed \cite{becker2007weakly} based on the Tait equation.

\paragraph{Inference by physical simulation} This SPH-based inference method falls into a class of physics-inspired, approximate inference methods which use physical process to generate samples, e.g. Hamiltonian Monte Carlo (HMC), Langevin dynamics \cite{Roberts1996}, electrostatics-based ParVI \cite{huang2024EParVI}, etc. This opens a new door for variational inference (VI): parameters are traced though a principled physical simulation whose intermediate or terminal state gives the target distribution. It is different from simulation-based inference (SBI) in that, inference by simulation (IBS) performs physical simulations over the inference parameters, while SBI (e.g. ABC) performs statistical simulations over a quantity given the parameters.

\paragraph{Error sources} The are two types of approximation errors in SPH \cite{sigalotti2021mathematics}: kernel approximation error and particle (discretization) approximation error. The former is independent of particle distribution and replies on the smoothing length $h$; the later originates from the discretization procedure and is a function of the number of neighbours within the kernel support which depends on the actual distribution of particles. \cite{zhu2015numerical} suggests that using small $h$ and large number of neighbors \cite{sigalotti2016kernel,Gabbasov2017consistent,Velasquez2018impetus} at the same time can reduce the overall error, improve SPH accuracy and convergence, and achieve full particle consistency in the limit \cite{Rasio2000particle}, at the cost of increasing computational burden.

\section{Conclusion and future work}

\paragraph{Conclusion} This paper presents the theoretical framework of a new physics-based sampling and variational inference method, \textit{SPH-ParVI}. This method formulates a sampling problem as a physical simulation process which configures a fluid from an initial shape (prior distribution) to a geometry that approximates the target distribution. The fluid is discretized as an interacting particle system (IPS) via the powerful SPH method: particles carry interactive, smoothed properties and their movements are dominantly driven by external effects. Two formulations of the SPH-ParVI method are proposed: incorporating the target density (if density is partially known) or gradient (if only gradient information are available) into the SPH simulation as an external pressure or force field. This modified SPH method, with governing physics inheriting the Navier-Stokes equations, provides fast, flexible, scalable and deterministic sampling and inference for complex (e.g. multi-modal) and high-dimensional densities.

\paragraph{Future work} Rather than re-inventing the physics, this preliminary work focuses on building a framework for applying the SPH method to the statistical regime; experimental validations are to be included in a future version. Convergence and potentially uniqueness of the steady state geometry after introducing external effects demand rigorous mathematical treatment. Approximation errors, e.g. hypothetical and numerical errors resulted from assumptions (e.g. linear viscosity) and discretization, require formal analysis. Constrained inference, e.g. inference on a manifold, can be realised by imposing boundary conditions. Probabilistic move can be inserted as an Metropolis–Hastings step. Other SPH variants\footnote{For example, the corrective smoothed particle method (CSPM), finite particle method (FPM), modified smoothed particle hydrodynamics (MSPH) have been invented to restore SPH consistency and improve accuracy at an increased computational cost and loss of numerical stability \cite{sigalotti2021mathematics}.} can be explored. Efficiency may be improved by using Koopman operator for approximating the dynamics.

\bibliography{reference}

\begin{thebibliography}{10}

\bibitem{sph2019recent}
Smoothed particle hydrodynamics (sph) for complex fluid flows: Recent developments in methodology and applications.
\newblock {\em Physics of Fluids}, 31(1):011301, 1 2019.

\bibitem{Antuono2010}
M.~Antuono, A.~Colagrossi, S.~Marrone, and D.~Molteni.
\newblock Free-surface flows solved by means of sph schemes with numerical diffusive terms.
\newblock {\em Computer Physics Communications}, 181:532--549, 2010.

\bibitem{becker2007weakly}
Markus Becker and Matthias Teschner.
\newblock Weakly compressible sph for free surface flows.
\newblock In D.~Metaxas and J.~Popovic, editors, {\em Eurographics/ ACM SIGGRAPH Symposium on Computer Animation}, pages 1--8. University of Freiburg, 2007.

\bibitem{bell2005particle}
N.~Bell, Y.~Yu, and P.~J. Mucha.
\newblock Particle-based simulation of granular materials.
\newblock In {\em SCA '05: Proceedings of the 2005 ACM SIGGRAPH/Eurographics Symposium on Computer Animation}, pages 77--86, New York, NY, USA, 2005. ACM Press.

\bibitem{bonet1999variational}
J.~Bonet and T.-S.L. Lok.
\newblock Variational and momentum preservation aspects of smooth particle hydrodynamic formulations.
\newblock {\em Computer Methods in Applied Mechanics and Engineering}, 180:97--115, 1999.

\bibitem{bridson2006fluid}
Robert Bridson, Ronald Fedkiw, and Matthias M{\"u}ller-Fischer.
\newblock Fluid simulation: Siggraph 2006 course notes.
\newblock In {\em SIGGRAPH '06: ACM SIGGRAPH 2006 Courses}, pages 1--87, New York, NY, USA, 2006. ACM Press.

\bibitem{chen1999completeness}
J.K. Chen, J.E. Beraun, and C.J. Jih.
\newblock Completeness of corrective smoothed particle method for linear elastodynamics.
\newblock {\em Computational Mechanics}, 24:273--285, 1999.

\bibitem{chen1999improvement}
J.K. Chen, J.E. Beraun, and C.J. Jih.
\newblock An improvement for tensile instability in smoothed particle hydrodynamics.
\newblock {\em Computational Mechanics}, 23:279--287, 1999.

\bibitem{clavet2005particle}
Simon Clavet, Philippe Beaudoin, and Pierre Poulin.
\newblock Particle-based viscoelastic fluid simulation.
\newblock In {\em SCA '05: Proceedings of the 2005 ACM SIGGRAPH/Eurographics Symposium on Computer Animation}, pages 219--228, New York, NY, USA, 2005. ACM Press.

\bibitem{desbrun1996smoothed}
Mathieu Desbrun and Marie-Paule Cani.
\newblock Smoothed particles: A new paradigm for animating highly deformable bodies.
\newblock In {\em 6th Eurographics Workshop on Computer Animation and Simulation '96}, pages 61--76, 1996.

\bibitem{Devroye1986}
Luc Devroye.
\newblock {\em Non-Uniform Random Variate Generation}.
\newblock Springer-Verlag, New York, 1986.

\bibitem{diBlasi2009consistency}
G.~Di~Blasi, E.~Francomano, A.~Tortorici, and E.~Toscano.
\newblock On the consistency restoring in sph.
\newblock In J.~Vigo-Aguiar, editor, {\em Proceedings of the International Conference on Computational and Mathematical Methods in Science and Engineering}, pages 393--404. CMMSE, 2009.

\bibitem{Doucet2001}
Arnaud Doucet, Nando de~Freitas, and Neil Gordon.
\newblock {\em Sequential Monte Carlo Methods in Practice}.
\newblock Springer, New York, 2001.

\bibitem{Duane1987}
Simon Duane, A.~D. Kennedy, B.~J. Pendleton, and D.~Roweth.
\newblock Hybrid monte carlo.
\newblock {\em Physics Letters B}, 195(2):216--222, 1987.

\bibitem{ferrand2013unified}
M.~Ferrand, D.R. Laurence, B.D. Rogers, D.~Violeau, and C.~Kassiotis.
\newblock Unified semi-analytical wall boundary conditions for inviscid, laminar or turbulent flows in the meshless sph method.
\newblock {\em International Journal for Numerical Methods in Fluids}, 71:446--472, 2013.

\bibitem{ferrand2011consistent}
M.~Ferrand, D.~Violeau, B.D. Rogers, and D.R.P. Laurence.
\newblock Consistent wall boundary treatment for laminar and turbulent flows in sph.
\newblock In {\em Proceedings of the 6th International SPHERIC SPH Workshop}, pages 275--282, 2011.

\bibitem{Bayesian_signal_processing_Joseph}
Joseph J.K. O Ruanaidh; William~J. Fitzgerald.
\newblock {\em Numerical Bayesian Methods Applied to Signal Processing}, page 244.
\newblock Springer, 1st edition, 1996.

\bibitem{Gabbasov2017consistent}
R.~Gabbasov, L.~Di~G. Sigalotti, F.~Cruz, J.~Klapp, and J.M. Ramírez-Velasquez.
\newblock Consistent sph simulations of protostellar collapse and fragmentation.
\newblock {\em The Astrophysical Journal}, 835:287, 2017.

\bibitem{galdi2011introduction}
G.P. Galdi.
\newblock {\em An Introduction to the Mathematical Theory of the Navier-Stokes Equations}.
\newblock Springer Monographs in Mathematics. Springer New York, NY, 2 edition, 2011.
\newblock Hardcover ISBN 978-0-387-09619-3, Softcover ISBN 978-1-4939-5017-1, eBook ISBN 978-0-387-09620-9.

\bibitem{Geman1984}
Stuart Geman and Donald Geman.
\newblock Stochastic relaxation, gibbs distributions, and the bayesian restoration of images.
\newblock {\em IEEE Transactions on Pattern Analysis and Machine Intelligence}, PAMI-6(6):721--741, 1984.

\bibitem{Genton2001kernel}
Marc~G. Genton.
\newblock Classes of kernels for machine learning: A statistics perspective.
\newblock {\em Journal of Machine Learning Research}, 2:299--312, 2001.

\bibitem{gibiansky2019computational}
Andrew Gibiansky.
\newblock Computational fluid dynamics, 2019.
\newblock Accessed: 2024-06-10.

\bibitem{Gingold1977}
Robert~A. Gingold and Joseph~J. Monaghan.
\newblock Smoothed particle hydrodynamics: theory and application to non-spherical stars.
\newblock {\em Monthly Notices of the Royal Astronomical Society}, 181(3):375--389, 1977.

\bibitem{han2018sph}
L.~Han and X.~Hu.
\newblock Sph modeling of fluid-structure interaction.
\newblock {\em Journal of Hydrodynamics}, 30:62--69, 2018.

\bibitem{Ho2020DDPM}
Jonathan Ho, Ajay Jain, and Pieter Abbeel.
\newblock Denoising diffusion probabilistic models.
\newblock {\em arXiv preprint arxiv:2006.11239}, 2020.

\bibitem{Hoffman2013}
Matthew~D. Hoffman, David~M. Blei, Chong Wang, and John Paisley.
\newblock Stochastic variational inference.
\newblock {\em Journal of Machine Learning Research}, 14(1):1303--1347, 2013.

\bibitem{huang2015kernel}
C.~Huang, J.M. Lei, M.B. Liu, and X.Y. Peng.
\newblock A kernel gradient free (kgf) sph method.
\newblock {\em International Journal for Numerical Methods in Fluids}, 78:691--707, 2015.

\bibitem{HUANG2019571}
C.~Huang, T.~Long, S.M. Li, and M.B. Liu.
\newblock A kernel gradient-free sph method with iterative particle shifting technology for modeling low-reynolds flows around airfoils.
\newblock {\em Engineering Analysis with Boundary Elements}, 106:571--587, 2019.

\bibitem{SBGM_Huang}
Yongchao Huang.
\newblock Classification via score-based generative modelling.
\newblock {\em https://arxiv.org/abs/2207.11091}, 2022.

\bibitem{huang2024EParVI}
Yongchao Huang.
\newblock Electrostatics-based particle sampling and approximate inference.
\newblock {\em https://arxiv.org/abs/2406.20044}, 2024.

\bibitem{hyvarinen2005sm}
Aapo Hyv{\"a}rinen.
\newblock Estimation of non-normalized statistical models by score matching.
\newblock {\em Journal of Machine Learning Research}, 6:695--709, 2005.

\bibitem{Jordan1999introduction}
Michael~I. Jordan, Zoubin Ghahramani, Tommi~S. Jaakkola, and Lawrence~K. Saul.
\newblock An introduction to variational methods for graphical models.
\newblock {\em Machine Learning}, 37:183--233, 1999.

\bibitem{Kahn1949}
Herman Kahn and Theodore~E. Harris.
\newblock Estimation of particle transmission by random sampling.
\newblock In {\em Monte Carlo Method}, volume~12 of {\em Applied Mathematics Series}, pages 27--30. National Bureau of Standards, 1949.

\bibitem{keiser2005unified}
Richard Keiser, Bryan Adams, Dominic Gasser, Pascal Bazzi, Philip Dutré, and Markus Gross.
\newblock A unified lagrangian approach to solid-fluid animation.
\newblock In {\em Eurographics Symposium on Point-Based Graphics}, pages 125--133, 2005.

\bibitem{kingma2013VAE}
Diederik~P Kingma and Max Welling.
\newblock Auto-encoding variational bayes, 2013.

\bibitem{lamb1994galactic}
S.A. Lamb, R.A. Gerber, and D.S. Balsara.
\newblock Galactic scale gas flows in colliding galaxies: 3-dimensional, n-body/hydrodynamics experiments.
\newblock {\em Astrophysics and Space Science}, 216:337--346, 1994.

\bibitem{landau1987fluid}
L.D. Landau and E.M. Lifshitz.
\newblock {\em Fluid Mechanics}, volume~6 of {\em Course of Theoretical Physics}.
\newblock Pergamon Press, Oxford, 1987.

\bibitem{li1996moving}
S.~Li and W.K. Liu.
\newblock Moving least-square reproducing kernel method part ii: Fourier analysis.
\newblock {\em Computer Methods in Applied Mechanics and Engineering}, 139:159--193, 1996.

\bibitem{Lind2020}
Steven~J. Lind, Benedict~D. Rogers, and Peter~K. Stansby.
\newblock Review of smoothed particle hydrodynamics: towards converged lagrangian flow modelling.
\newblock {\em Proceedings of the Royal Society A}, 476(20190801), 2020.

\bibitem{litvinov2015towards}
S.~Litvinov, X.Y. Hu, and N.A. Adams.
\newblock Towards consistence and convergence of conservative sph approximations.
\newblock {\em Journal of Computational Physics}, 301:394--401, 2015.

\bibitem{liu2006restoring}
M.B. Liu and G.R. Liu.
\newblock Restoring particle consistency in smoothed particle hydrodynamics.
\newblock {\em Applied Numerical Mathematics}, 56:19--36, 2006.

\bibitem{liu2003constructing}
M.B. Liu, G.R. Liu, and K.Y. Lam.
\newblock Constructing smoothing functions in smoothed particle hydrodynamics with applications.
\newblock {\em Journal of Computational and Applied Mathematics}, 155:263--284, 2003.

\bibitem{Liu2016SVGD}
Qiang Liu and Dilin Wang.
\newblock Stein variational gradient descent: a general purpose bayesian inference algorithm.
\newblock In {\em Proceedings of the 30th International Conference on Neural Information Processing Systems}, NIPS'16, page 2378–2386, Red Hook, NY, USA, 2016. Curran Associates Inc.

\bibitem{liu1997moving}
W.-K. Liu, S.~Li, and T.~Belytschko.
\newblock Moving least-square reproducing kernel methods (i) methodology and convergence.
\newblock {\em Computer Methods in Applied Mechanics and Engineering}, 143:113--154, 1997.

\bibitem{Lucy1977numerical}
L.B. Lucy.
\newblock A numerical approach to the testing of the fission hypothesis.
\newblock {\em The Astronomical Journal}, 82:1013--1024, 1977.

\bibitem{Max1981vectorized}
N.~Max.
\newblock Vectorized procedural models for natural terrain: Waves and islands in the sunset.
\newblock In {\em SIGGRAPH '81: Proceedings of the 8th Annual Conference on Computer Graphics and Interactive Techniques}, pages 317--324, New York, NY, USA, 1981. ACM Press.

\bibitem{Metropolis1953}
Nicholas Metropolis, Arianna~W. Rosenbluth, Marshall~N. Rosenbluth, Augusta~H. Teller, and Edward Teller.
\newblock Equation of state calculations by fast computing machines.
\newblock {\em The Journal of Chemical Physics}, 21(6):1087--1092, 1953.

\bibitem{miller1989globular}
Gavin Miller and Alan Pearce.
\newblock Globular dynamics: A connected particle system for animating viscous fluids.
\newblock {\em Computers and Graphics}, 13(3):305--309, 1989.

\bibitem{mocz2011smoothed}
Philip Mocz.
\newblock Smoothed particle hydrodynamics: Theory, implementation, and application to toy stars, 2011.

\bibitem{monaghan1994simulating}
J.~J. Monaghan.
\newblock Simulating free surface flows with sph.
\newblock {\em Journal of Computational Physics}, 110(2):399--406, 1994.

\bibitem{Monaghan2005}
J.~J. Monaghan.
\newblock Smoothed particle hydrodynamics.
\newblock {\em Reports on Progress in Physics}, 68(8):1703, 2005.

\bibitem{Morris1999surface}
J.~Morris.
\newblock Simulating surface tension with smoothed particle hydrodynamics.
\newblock {\em International Journal for Numerical Methods in Fluids}, 33:333--353, 1999.

\bibitem{Morrison2012SubstantialDerivative}
Frank~C. Morrison.
\newblock Substantial derivative.
\newblock \url{https://pages.mtu.edu/~fmorriso/cm4650/2012SubstantialDerivative.pdf}, 2012.
\newblock Accessed: 2024-07-04.

\bibitem{muller2003particle}
Matthias Müller, David Charypar, and Markus Gross.
\newblock Particle-based fluid simulation for interactive applications.
\newblock In {\em Proceedings of the 2003 ACM SIGGRAPH/Eurographics Symposium on Computer Animation}, pages 154--159, Aire-la-Ville, Switzerland, Switzerland, 2003. Eurographics Association.
\newblock SCA '03.

\bibitem{muller2004interaction}
Matthias Müller, Stefan Schirm, Matthias Teschner, Bruno Heidelberger, and Markus Gross.
\newblock Interaction of fluids with deformable solids.
\newblock {\em Computer Animation and Virtual Worlds}, 15(3-4):159--171, 2004.

\bibitem{nasreldeen2017sph}
Ahmed Nasreldeen, Long Fan, and Keru Liu.
\newblock Smooth particle hydrodynamics (sph).
\newblock Technical report, Penn State University, College of Earth and Mineral Sciences, Department of Energy and Mineral Engineering, 2017.

\bibitem{Neal2003}
Radford~M. Neal.
\newblock Slice sampling.
\newblock {\em Annals of Statistics}, 31(3):705--767, 6 2003.

\bibitem{Niederreiter1992}
Harald Niederreiter.
\newblock {\em Random Number Generation and Quasi-Monte Carlo Methods}, volume~63 of {\em CBMS-NSF Regional Conference Series in Applied Mathematics}.
\newblock SIAM, Philadelphia, 1992.

\bibitem{obrien1995dynamic}
James~F. O'Brien and Jessica~K. Hodgins.
\newblock Dynamic simulation of splashing fluids.
\newblock In {\em CA '95: Proceedings of the Computer Animation}, page 198, Washington, DC, USA, 1995. IEEE Computer Society.

\bibitem{Peachey1986modeling}
D.~Peachey.
\newblock Modeling waves and surf.
\newblock In {\em SIGGRAPH '86: Proceedings of the 13th Annual Conference on Computer Graphics and Interactive Techniques}, pages 65--74, New York, NY, USA, 1986. ACM Press.

\bibitem{premoze2003particle}
Simon Premoze, Tolga Tasdizen, James Bigler, Aaron Lefohn, and Ross Whitaker.
\newblock Particle-based simulation of fluids.
\newblock {\em Computer Graphics Forum (Proc. of Eurographics)}, 22:401--410, 2003.

\bibitem{Velasquez2018impetus}
J.M. Ramírez-Velasquez, L.~Di~G. Sigalotti, R.~Gabbasov, F.~Cruz, and J.~Klapp.
\newblock Impetus: Consistent sph calculations of 3d spherical bondi accretion on to a black hole.
\newblock {\em Monthly Notices of the Royal Astronomical Society}, 477:4308--4329, 2018.

\bibitem{Ranganath2014}
Rajesh Ranganath, Sean Gerrish, and David~M. Blei.
\newblock Black box variational inference.
\newblock In {\em Proceedings of the 17th International Conference on Artificial Intelligence and Statistics (AISTATS)}, pages 814--822, 2014.

\bibitem{Rasio2000particle}
F.A. Rasio.
\newblock Particle methods in astrophysical fluid dynamics.
\newblock {\em Progress of Theoretical Physics Supplement}, 138:609--621, 2000.

\bibitem{reinhardt2019consistent}
S.~Reinhardt, T.~Krake, B.~Eberhardt, and D.~Weiskopf.
\newblock Consistent shepard interpolation for sph-based fluid animation.
\newblock {\em ACM Transactions on Graphics}, 38:1--11, 2019.

\bibitem{Roberts1996}
Gareth~O. Roberts and Richard~L. Tweedie.
\newblock Exponential convergence of langevin distributions and their discrete approximations.
\newblock {\em Bernoulli}, 2(4):341--363, 1996.

\bibitem{Roberts2002AR}
Stephen~J Roberts and Will~D Penny.
\newblock Variational bayes for generalized autoregressive models.
\newblock {\em IEEE Transactions on Signal Processing}, 50(9):2245--2257, 2002.

\bibitem{Roselli201920}
Riccardo Angelini~Rota Roselli, Giuliano Vernengo, Stefano Brizzolara, and Roberto Guercio.
\newblock Sph simulation of periodic wave breaking in the surf zone - a detailed fluid dynamic validation.
\newblock {\em Ocean Engineering}, 176:20--30, 2019.

\bibitem{Shadloo2016}
M.~S. Shadloo, G.~Oger, and D.~L. Touze.
\newblock Smoothed particle hydrodynamics method for fluid flows, towards industrial applications: Motivations, current state, and challenges.
\newblock {\em Computers and Fluids}, 136:11--34, 2016.

\bibitem{sibilla2015algorithm}
S.~Sibilla.
\newblock An algorithm to improve consistency in smoothed particle hydrodynamics.
\newblock {\em Computers \& Fluids}, 118:148--158, 2015.

\bibitem{sigalotti2016kernel}
L.~Di~G. Sigalotti, J.~Klapp, O.~Rendón, C.A. Vargas, and F.~Peña-Polo.
\newblock On the kernel and particle consistency in smoothed particle hydrodynamics.
\newblock {\em Applied Numerical Mathematics}, 108:242--255, 2016.

\bibitem{sigalotti2021mathematics}
Leonardo Di~G. Sigalotti, Jaime Klapp, and Moncho~Gómez Gesteira.
\newblock The mathematics of smoothed particle hydrodynamics (sph) consistency.
\newblock {\em Frontiers in Applied Mathematics and Statistics}, 7, 2021.

\bibitem{SimScaleNavierStokes}
SimScale.
\newblock What are the navier-stokes equations?
\newblock \url{https://www.simscale.com/docs/simwiki/numerics-background/what-are-the-navier-stokes-equations/}.
\newblock Accessed: 2024-07-04.

\bibitem{scoreSDE2020}
Yang Song, Jascha Sohl{-}Dickstein, Diederik~P. Kingma, Abhishek Kumar, Stefano Ermon, and Ben Poole.
\newblock Score-based generative modeling through stochastic differential equations.
\newblock {\em CoRR}, abs/2011.13456, 2020.

\bibitem{stam1995fire}
Jos Stam and Eugene Fiume.
\newblock Depicting fire and other gaseous phenomena using diffusion processes.
\newblock In {\em Proceedings of the 22nd Annual Conference on Computer Graphics and Interactive Techniques}, SIGGRAPH '95, pages 129--136, New York, NY, USA, 1995. ACM Press.

\bibitem{stora1999animating}
D.~Stora, P.~Agliati, M.-P. Cani, F.~Neyret, and J.~Gascuel.
\newblock Animating lava flows.
\newblock In {\em Graphics Interface 99}, pages 203--210, 1999.

\bibitem{Damien2012}
Damien Violeau.
\newblock {\em {Fluid Mechanics and the SPH Method: Theory and Applications}}.
\newblock Oxford University Press, 05 2012.

\bibitem{vonNeumann1951}
John von Neumann.
\newblock Various techniques used in connection with random digits.
\newblock {\em Applied Math Series, Notes by G. E. Forsythe, in National Bureau of Standards}, 12:36--38, 1951.

\bibitem{Wang2021EVI}
Yiwei Wang, Jiuhai Chen, Chun Liu, and Lulu Kang.
\newblock Particle-based energetic variational inference.
\newblock {\em Statistics and Computing}, 31(34), 2021.

\bibitem{White2002fluid}
Frank White.
\newblock {\em Fluid Mechanics}.
\newblock McGraw-Hill Higher Education, 4th edition, 2002.

\bibitem{Xu2021}
Xiaoyang Xu, Yao-Lin Jiang, and Peng Yu.
\newblock Sph simulations of 3d dam-break flow against various forms of the obstacle: Toward an optimal design.
\newblock {\em Ocean Engineering}, 229:108978, 2021.

\bibitem{zhang2004modified}
G.M. Zhang and R.C. Batra.
\newblock Modified smoothed particle hydrodynamics method and its application to transient problems.
\newblock {\em Computational Mechanics}, 34:137--146, 2004.

\bibitem{zhou2008accuracy}
D.~Zhou, S.~Chen, L.~Li, H.~Li, and Y.~Zhao.
\newblock Accuracy improvement of smoothed particle hydrodynamics.
\newblock {\em Engineering Applications of Computational Fluid Mechanics}, 2:244--251, 2008.

\bibitem{zhu2015numerical}
Q.~Zhu, L.~Hernquist, and Y.~Li.
\newblock Numerical convergence in smoothed particle hydrodynamics.
\newblock {\em The Astrophysical Journal}, 800(1):6, 2015.

\end{thebibliography}

\appendix

\section{Mathematical notations} \label{app:mathematical_notations}

For clarity, we present some of the (common) mathematical notations used in this context.

\paragraph{Gradient} Gradient is a collection of derivatives along each coordinate. For a real-valued function \( f(x, y, z) \) on \( \mathbb{R}^3 \), the gradient \( \nabla f(x, y, z) \) is a vector-valued function on \( \mathbb{R}^3 \):

\[
\nabla f(x, y, z) = \left( \frac{\partial f}{\partial x}, \frac{\partial f}{\partial y}, \frac{\partial f}{\partial z} \right) = \frac{\partial f}{\partial x} \mathbf{i} + \frac{\partial f}{\partial y} \mathbf{j} + \frac{\partial f}{\partial z} \mathbf{k}
\]

\paragraph{Divergence} The divergence \(\text{div} \, \mathbf{f}\) is the dot product of \(\mathbf{f}\) with \(\nabla\) (the gradient operator can be regarded as a vector):
\[
\text{div} \, \mathbf{f} = \left( \frac{\partial}{\partial x} \mathbf{i} + \frac{\partial}{\partial y} \mathbf{j} + \frac{\partial}{\partial z} \mathbf{k} \right) \cdot [ f_1(x, y, z) \mathbf{i} + f_2(x, y, z) \mathbf{j} + f_3(x, y, z) \mathbf{k} ]
\]
\[
= \left( \frac{\partial}{\partial x} \right) (f_1) + \left( \frac{\partial}{\partial y} \right) (f_2) + \left( \frac{\partial}{\partial z} \right) (f_3)
= \frac{\partial f_1}{\partial x} + \frac{\partial f_2}{\partial y} + \frac{\partial f_3}{\partial z}
\]

\noindent Often, it is convenient to write the divergence \(\text{div} \, \mathbf{f}\) as \(\nabla \cdot \mathbf{f}\).

\paragraph{Laplacian} For a real-valued function \( f(x, y, z) \), the \textit{Laplacian} of \( f \), denoted by \( \Delta f \), is:
\[
\Delta f(x, y, z) = \nabla \cdot \nabla f = \frac{\partial^2 f}{\partial x^2} + \frac{\partial^2 f}{\partial y^2} + \frac{\partial^2 f}{\partial z^2}.
\]

\noindent Often, instead of using \( \Delta f \), the notation \( \nabla^2 f \) is used for the Laplacian, with the convention \( \nabla^2 = \nabla \cdot \nabla \).

In this context, derivative(s) in $\nabla$, $\nabla \cdot$ and $\nabla^2$ all refer to spatial derivative(s), i.e. \textit{w.r.t }the coordinates. $\nabla K_{ij}$, for example, is an abbreviation for $\nabla_i K_{ij} = \nabla_{\textbf{r}} K(|\textbf{r} - \textbf{r}_j|,h) |_{\textbf{r}=\textbf{r}_i}$.

\section{Computational fluid dynamics: basics} 

Fluid dynamics is a sub-field of fluid mechanics which concerns about the flow of fluids such as liquids (e.g. hydrodynamics) and gases (e.g. aerodynamics). There are typically two classes of methods used in computational fluid dynamics (CFD): \textit{discretization models} and \textit{turbulence models} \cite{gibiansky2019computational}. The former divides the fluid into a set of discrete elements (e.g. particles), and iteratively simulate their motions using the Navier-Stokes equations; turbulence models attempt to track or obtain the value of specific properties of a fluid over time. Particle-based methods simplify computation and offer approximate, interactive simulations with fast speed: each particle is subject to the velocity fields dictated by the Navier-Stokes equations as well as external forces.

In the center of fluid dynamics are the Navier-Stokes (N-S) equations, which are partial differential equations that describe the motion of viscous fluids \cite{galdi2011introduction}. N-S equations are based Newton’s second law of motion, taking into account the viscous and pressure effects in order to describe a viscous fluid flow. These equations consist of three types of conservation equations: conservation equations of mass (\textit{continuity equation} \footnote{Continuity equations are a stronger, local form of conservation laws, which states that the quantity is locally conserved: it can neither be created nor destroyed, nor can it "teleport" from one place to another. It can only move by a continuous flow. A weak version of the conservation law states that, the quantity could disappear from one point while simultaneously appearing at another point, with the total amount fixed.}), momentum (\textit{Newton's second law} for fluids), and energy (\textit{first law of thermodynamics}). \textbf{Conservation of mass} states that the net difference between mass inflow and outflow throughout the fluid is zero:

\begin{equation} \label{eq:continuity_equation_general}
    \frac{\partial \rho}{\partial t} + \nabla \cdot (\rho \mathbf{v}) = 0 \quad \text{(Continuity equation)}
\end{equation}

\noindent where $\rho$ is density, $\mathbf{v}$ is velocity. The term $\nabla \cdot (\rho \mathbf{v})$ captures the difference between the inflow and outflow of mass at any point in the fluid. As $\rho$ is a scalar-valued function and $\mathbf{v}$ is a vector field, as per the properties of divergence, we have \footnote{The dot product $(\nabla \rho) \cdot \mathbf{v}=(\nabla \rho)^T \mathbf{v}$ gives a scalar, the divergence $\nabla \cdot \mathbf{v}$ is also a scalar.} $\nabla \cdot (\rho \mathbf{v}) = (\nabla \rho) \cdot \mathbf{v} + \rho (\nabla \cdot \mathbf{v})$, which expands Eq.\ref{eq:continuity_equation_general} as:

\[
\frac{\partial \rho}{\partial t} + (\nabla \rho) \cdot \mathbf{v} + \rho (\nabla \cdot \mathbf{v}) = 0
\]

\noindent the first two terms together give the material derivative $\frac{D \rho}{D t} = \frac{\partial \rho}{\partial t} + \mathbf{v} \cdot \nabla \rho$, which simplifies the above equation to

\[
\frac{D \rho}{D t} + \rho \nabla \cdot \mathbf{v}= 0.
\]

\noindent If the fluid is \textit{incompressible}, we have $\frac{D \rho}{D t}=0$, which simplifies the continuity equation (three-dimensional fluid as an example):

\[
\frac{D \rho}{D t} = 0 \rightarrow \nabla \cdot \mathbf{v} = \frac{\partial \mathbf{v}}{\partial x} + \frac{\partial \mathbf{v}}{\partial y} + \frac{\partial \mathbf{v}}{\partial z} = 0
\quad \text{(\textbf{Incompressible} continuity equation)}
\]

The other type of fluid is \textit{compressible fluids} \footnote{Although all types of fluids are compressible over a variety of molecular structures, most of them, however, can be assumed incompressible \cite{SimScaleNavierStokes}, i.e. their density changes are negligible. There are exceptional scenarios where fluids should be considered as compressible, one is geophysical flows in which thermal effects can lead to density changes, another is high-speed flows in which the velocity exceeds a critical limit (characterized by the \textit{Mach number} threshold).} which transport the density as per the mass conservation law Eq.\ref{eq:continuity_equation_general}. Representing a fluid using a set of particles naturally preserves mass: conservation of mass is guaranteed for particle systems as the number of particles are fixed with constant mass.

Momentum is the produce of mass and velocity, describing a mass in motion. According to Newton's second law, $\textbf{f}=m \textbf{a}=m \Delta \mathbf{v} / \Delta t$ where $f$ is the net force, $m$ the mass, $\textbf{a}$ the acceleration, the momentum can only change through the action of forces; \textbf{conservation of momentum} (i.e. momentum remains constant) indicates momentum is neither created nor destroyed. The Newton's second law for a fluid can be expressed as \cite{SimScaleNavierStokes}:

\[
\rho \frac{D \mathbf{v}}{D t} = \textbf{f} = \textbf{f}^{body} + \textbf{f}^{surf}
\]

\noindent where $\frac{D}{Dt}$, with capital $D$, is the \textit{substantial derivative} \footnote{The substantial derivative of $Q$, in 3D, is: $\frac{DQ}{Dt}={\textstyle {\frac {\partial Q}{\partial t}}+\mathbf{v} \cdot \nabla Q}=\frac{\partial Q}{\partial t} + \frac{\partial Q}{\partial x} v_x + \frac{\partial Q}{\partial y} v_y + \frac{\partial Q}{\partial z} v_z$. The substantial derivative represents the rate of change of the quantity $Q$ ($Q$ can be mass, energy, momentum, etc) as experienced by an observer that is moving with the flow (the Lagrangian approach). The observations made by a moving observer are affected by the stationary time-rate-of-change $\frac{\partial Q}{\partial t}$, as well as where the observer goes as it floats along with the flow ($\mathbf{v} \cdot \nabla Q$) \cite{Morrison2012SubstantialDerivative}.} (also termed \textit{material derivative}). $\textbf{f}$ is the net force exerted on a fluid element (e.g. a particle) per unit volume. $\textbf{f}^{body}$ is the body force \footnote{A body force is a force that acts throughout the volume of a body (e.g. a particle or a material), e.g. forces induced by friction, gravity, electric fields and magnetic fields, etc. Body forces contrast with contact forces or surface forces which are exerted to the surface of a body. In the context of fluid dynamics and the Navier-Stokes equations, body forces are forces that act on every element of the fluid within a volume. These forces distribute over the volume of the fluid rather than being applied at specific points or surfaces.} applied on the whole volume of the particle, e.g. gravitational force $\textbf{f}^{body}=\rho \textbf{g}$ where $\textbf{g}$ is the gravitational acceleration. In this context, we denote all external body force per unit volume as $\textbf{f}^{body}=\textbf{f}^{ext}$. $\textbf{f}^{surf}$ is the internal, surface force such as those induced by pressure (pressure gradient force $\textbf{f}^{pre}$) and viscosity (viscous force $\textbf{f}^{vis}$), it can be in general expressed as:
\[
\textbf{f}^{surf} = \textbf{f}^{pre} + \textbf{f}^{vis} = \nabla \cdot \tau
\]
\noindent where $\tau$ is a stress tensor which can be decomposed into pressure gradient stress and viscous stress \footnote{Viscous stress can be further decomposed into those induced by velocity gradients and bulk viscosity, see e.g. \cite{SimScaleNavierStokes}.}. $\nabla \cdot \tau$ is the (spatial) divergence of the stress tensor, indicating how the stress varies in space. Bringing the body and surface forces together, we re-write the Newton's second law for fluid as:

\begin{equation} \label{eq:momentum_equation_general} 
    \rho \frac{D \mathbf{v}}{D t} = \textbf{f} = \textbf{f}^{ext} + \textbf{f}^{pre} + \textbf{f}^{vis}
\end{equation}

The LHS of the momentum equation represents inertia force (i.e. acceleration of the fluid element); RHS are contributions from body forces, pressure gradients, and internal friction. There are cases where the viscous force $\textbf{f}^{vis}$ is negligible (i.e. \textit{inviscid flow}, and the N-S momentum equation reduces to the \textit{Euler equation} \cite{becker2007weakly}), e.g. when the \textit{Reynolds number}($Re$), which is the ratio of inertial and viscous effects, approaches infinity ($Re \rightarrow \infty$). On the other hand, if $Re \ll 1$, then inertia effects are negligible (i.e. LHS becomes zero).

Incompressible fluids, again for example, have constant density and therefore the following N-S equations \footnote{ $\nabla$ denotes gradient (\textit{w.r.t.} coordinates), $\nabla \cdot$ denotes \textit{divergence}, $\nabla^2$ indicates \textit{Laplacian}. See Appendix.\ref{app:mathematical_notations}.}:

\begin{equation} \label{eq:NS_equations_incompressible1}
\begin{aligned}
    & \nabla \cdot \mathbf{v} = 0 \quad \text{(\textbf{Incompressibe} continuity equation)} \\
    & \rho \frac{D \mathbf{v}}{D t} = - \nabla P + \mu \nabla^2 \mathbf{v} + \mathbf{f}^{ext} \quad \text{(\textbf{Incompressible} momentum equation)} \\
\end{aligned}
\end{equation}

\noindent where \\
- $\mathbf{v}$ is the flow velocity vector. \\
- \(\rho\) is the density, i.e. the mass per unit volume of the fluid. It is a scalar field that varies with position and time. \\
- $\frac{D}{Dt}$ is the material derivative, defined as ${\textstyle {\frac {\partial }{\partial t}}+\mathbf{v} \cdot \nabla }$. \\
- $\nabla \cdot$ is the divergence, e.g. $\nabla \cdot \mathbf{v}$ is the divergence of the flow velocity $\mathbf{v}$, which is a scalar field. \\
- \(P\) is the pressure, a scalar field representing the force exerted per unit area within the fluid. It varies with position and time. $\nabla P$ is the gradient of the pressure field $P$, which is a vector \footnote{The gradient of a scalar field results in a vector field that points in the direction of the greatest rate of increase of the scalar. By default we assume column vectors.}. \\
- \(\mu\) is the \textit{dynamic viscosity} or \textit{absolute viscosity}. Here we have represented the viscous force as $\mu \nabla^2 \mathbf{v}$ where $\nabla^2 \mathbf{v}$ is the Laplacian of the velocity vector. \\
- \(\mathbf{f}^{ext}\) is the external (body) force per unit mass, e.g. gravitational force, electromagnetic forces, etc. \\

\noindent By defining the \textit{kinematic viscosity} or \textit{momentum diffusivity} $\nu = \mu / \rho$, Eq.\ref{eq:NS_equations_incompressible1} can be written in the more frequent form:

\begin{equation}  \label{eq:NS_equations_incompressible2} \tag{\ref{eq:NS_equations_incompressible1}b}
\begin{aligned}
    & \nabla \cdot \mathbf{v} = 0 \quad \text{(\textbf{Incompressibe} continuity equation)} \\
    & \frac{D \mathbf{v}}{D t} = - \frac{1}{\rho} \nabla P + \nu \nabla^2 \mathbf{v} + \frac{1}{\rho} \mathbf{f}^{ext} \quad \text{(\textbf{Incompressible} momentum equation)} \\
\end{aligned}
\end{equation}

\noindent the first term $- \frac{1}{\rho} \nabla P$ represents the force induced by spatial pressure gradient, which is caused by change of density, and is therefore an internal source. The pressure gradient drives the fluid flow from regions of high pressure to low pressure. The second term $\nu \nabla^2 \mathbf{v}$ represents viscous force within the fluid, which accounts for diffusion of momentum. $- \frac{1}{\rho} \nabla P$ and $\nu \nabla^2 \mathbf{v}$ together represent the spatial divergence of stress. $\mathbf{f}^{ext}$ represents contributions from external sources.

The \textbf{energy equation}, as per the first law of thermodynamics, has to be considered if any thermal interaction is involved in the problem. It specifies the total energy increment in the system as the sum of heat added to and work done on the system. As we are not concerned about thermal effects, we mainly focus on the mass and momentum conservation equations.

Also relevant \footnote{In the context of the Navier-Stokes equations, the EoS is essential for closing the system of equations, particularly for compressible flows.} is the \textbf{equation of state} (EoS), which relates pressure, temperature and density of a fluid \cite{landau1987fluid}:

\begin{equation} \label{eq:EoS_general}
    P = P(\rho, T)
\end{equation}

\noindent and indicates that pressure \(P\) is a function of density \( \rho \) and temperature \( T \). There exist various forms to relate pressure and density \cite{becker2007weakly}, one being the Poisson equation:

\begin{equation}
    \nabla^2 P = \rho \frac{\nabla \mathbf{v}}{\Delta t}
\end{equation}

\noindent which can be solved using Eulerian or Lagrangian approaches. For ideal gas \footnote{The \textit{ideal gas law} states $PV=nRT$, where $P$, $V$ and $T$ are the pressure, volume and temperature respectively. $n$ is the amount of substance, and $R$ is the ideal gas constant.}, we have \cite{muller2003particle,becker2007weakly,gibiansky2019computational}:

\begin{equation} \label{eq:EoS_ideal_gas1}
    P = k \rho
\end{equation}

\noindent or a more numerically stable version \cite{nasreldeen2017sph}:

\begin{equation} \label{eq:EoS_ideal_gas2} \tag{\ref{eq:EoS_ideal_gas1}b}
    P = c_s^2 (\rho - \rho_0)
\end{equation}

\noindent where $c_s$ is the sound speed, and $\rho_0$ is the rest (reference) density. Using either Eq.\ref{eq:EoS_ideal_gas1} or Eq.\ref{eq:EoS_ideal_gas2} is equivalent as the pressure gradient $\nabla P$ is concerned in the momentum equations.

The N-S equations have a complex, non-linear structure, and it is hardly possible to find an exact solution \cite{SimScaleNavierStokes}. Analytical solutions can be sought if assumptions are imposed to simplify the equations \cite{White2002fluid} (e.g. ignoring the nonlinearities, resulting in \textit{Couette flow}, \textit{Poisellie flow}, etc); numerical solutions via discretization can be obtained with different levels of complexity and accuracy. Solutions for the two-dimensional Navier-Stokes equations exist and are smooth; for low initial velocities, smooth solutions to 3-dimensional Navier-Stokes exist \cite{gibiansky2019computational}. 

\paragraph{Time-invariant N-S equations} If the flow is steady, i.e. the motion and parameters are independent of time, the term $\partial/\partial t$ vanishes, the general continuity (Eq.\ref{eq:continuity_equation_general}) and momentum (Eq.\ref{eq:momentum_equation_general}) equations are then simplified as:

\begin{equation} \label{eq:NS_equations_general_steady} 
\left\{
\begin{aligned}
    & \nabla \cdot (\rho \mathbf{v}) = 0 \quad \text{(Continuity equation)} \\
    & \rho \mathbf{v} \cdot \nabla \mathbf{v} = \textbf{f}^{ext} + \textbf{f}^{pre} + \textbf{f}^{vis} \quad \text{(Momentum equation)} \\
\end{aligned}
\right.
\end{equation}

\noindent the steady 3D N-S equations for incompressible fluids, for example, become:

\begin{equation} \label{eq:NS_equations_incompressible_steady} 
\left\{
\begin{aligned}
    & \frac{\partial v_x}{\partial x} + \frac{\partial v_y}{\partial y} + \frac{\partial v_z}{\partial z} = 0 \\
    & v_x \frac{\partial v_x}{\partial x} + v_y \frac{\partial v_x}{\partial y} + v_z \frac{\partial v_x}{\partial z} = - \frac{1}{\rho} \frac{\partial P}{\partial x} + \nu (\frac{\partial^2 v_x}{\partial x^2} + \frac{\partial^2 v_x}{\partial y^2} + \frac{\partial^2 v_x}{\partial z^2}) + \frac{1}{\rho} f^{ext}_x \\
    & v_x \frac{\partial v_y}{\partial x} + v_y \frac{\partial v_y}{\partial y} + v_z \frac{\partial v_y}{\partial z} = - \frac{1}{\rho} \frac{\partial P}{\partial y} + \nu (\frac{\partial^2 v_y}{\partial x^2} + \frac{\partial^2 v_y}{\partial y^2} + \frac{\partial^2 v_y}{\partial z^2}) + \frac{1}{\rho} f^{ext}_y \\
    & v_x \frac{\partial v_z}{\partial x} + v_y \frac{\partial v_z}{\partial y} + v_z \frac{\partial v_z}{\partial z} = - \frac{1}{\rho} \frac{\partial P}{\partial z} + \nu (\frac{\partial^2 v_z}{\partial x^2} + \frac{\partial^2 v_z}{\partial y^2} + \frac{\partial^2 v_z}{\partial z^2}) + \frac{1}{\rho} f^{ext}_z \\
\end{aligned}
\right.
\end{equation}

\section{SPH: fundamental principles} \label{app:SPH}

Here we follow \cite{Monaghan2005,Lind2020,Damien2012} and present the basic formulation of the SPH methodology.

For each particle $j$, as per Newton's second law, we have the acceleration:

\[
a_j = \frac{\mathbf{f}_j}{\rho_j}
\]

\noindent where $\mathbf{f}_j$ is the net force, i.e. sum of all forces acting on the particle.

\paragraph{SPH integral approximation}
A function $f(\textbf{r})$, which represents a physical quantity such as pressure or density at position $r$ in $d$-dimensional space $\Omega$, can be approximated via kernel approximation:

\begin{equation} \label{eq:kernel_approx}
    f(\textbf{r}) \approx \int_{\Omega} f(\textbf{r}') K(|\textbf{r}-\textbf{r}'|,h) dV(\textbf{r}')
\end{equation}

\noindent where $V$ is the volume of the particle. $h$ is defined as the smoothing length which characterizes the influence radius of the kernel $K(r,h)$. The kernel can be seen as a distance-based weight function. Normally it is chosen to be a smooth, isotropic and even function with compact support \cite{Lind2020}. General requirements of the smoothing kernel include \cite{sigalotti2021mathematics}: (1) positive definite; (2) even, i.e. $K(|\textbf{r} - \textbf{r}'|,h)=K(|\textbf{r}' - \textbf{r}|,h)$; (3) monotonically decreasing, i.e. the function value should decrease as distance increases (essentially decays to 0 at distances equal to or greater than multiples of $h$), as we expect nearby points should have more influence than those far away; (4) normalized, i.e. $\int K(|\textbf{r}|,h) d\textbf{r} = 1$; (5) it should equal the Dirac delta distribution in the limit $h \rightarrow 0$.

Commonly used kernel functions are Gaussian kernel and spline kernel. With positive kernels, the error of the integral approximation in Eq.\ref{eq:kernel_approx} is $\mathcal{O}(h^2)$ \cite{Monaghan2005}. Eq.\ref{eq:kernel_approx} gives the continuum approximation of the quantity $f(\textbf{r})$; the RHS integration in Eq.\ref{eq:kernel_approx} can be further approximated using discrete Riemann summation \footnote{Note the similarity between the SPH representation and spatial kernel smoother: both use neighbours to estimate the quantity at observation position. Kernel smoothing is used in e.g. kernel density estimation (KDE), kernel regression, etc.} \cite{sigalotti2021mathematics}:

\begin{equation} \label{eq:integral_discrete_approx}
    f(\textbf{r}) = \sum_{j=1}^M V_j f(\textbf{r}_j) K(|\textbf{r}-\textbf{r}_j|, h) = \sum_{j=1}^M \frac{m_j}{\rho_j} f(\textbf{r}_j) K(|\textbf{r}-\textbf{r}_j|, h)
\end{equation}

\noindent where $M$ is the total number of particles, $V_j=m_j/\rho_j$ is the volume of the $d$-dimensional particle $j$. Eq.\ref{eq:integral_discrete_approx} expresses the target quantity at position $\textbf{r}$ as a weighted summation of contributions from all the particles around the point. The discretization produces errors which reply on the particle size $V_j^{1/d}$, the influence length-scale $h$, and particle arrangement in space. 

The SPH formulation in Eq.\ref{eq:integral_discrete_approx} facilitates the calculation of derivatives of $f(\textbf{r})$ (spatial gradient, \textit{w.r.t.} coordinates $\textbf{r}$) and therefore discretization of the Navier-Stokes equations - we only need to differentiate the smoothing kernel. For example, 

\begin{equation} \label{eq:diff_SPH}
\begin{aligned}
&\nabla f(\textbf{r})= \frac{\partial}{\partial \textbf{r}} f(\textbf{r}) = \sum_{j=1}^M V_j f(\textbf{r}_j) \nabla K(|\textbf{r}-\textbf{r}_j|, h) \\
&\nabla^2 f(\textbf{r})= \frac{\partial^2}{\partial \textbf{r}^2} f(\textbf{r}) = \sum_{j=1}^M V_j f(\textbf{r}_j) \nabla^2 K(|\textbf{r}-\textbf{r}_j|, h) \\
\end{aligned}
\end{equation}

\paragraph{Density approximation} Replacing $f(\textbf{r}_j)$ by $\rho_j=\rho(\textbf{r}_j)$ in Eq.\ref{eq:integral_discrete_approx}, we obtain the density $\rho_i=\rho(\textbf{r}_i)$ of particle $i$ as:

\begin{equation} \label{eq:particle_density}
    \rho(\textbf{r}_i)=\sum_{j=1}^M V_j \rho_j K_{ij} =\sum_{j=1}^M m_j K_{ij}
\end{equation}

\noindent where for brevity we have denoted $K(|\textbf{r}_i-\textbf{r}_j|,h)$ as $K_{ij}$. Eq.\ref{eq:particle_density} gives erroneous (lower) density near surface for equally spaced particles, as surface particles in free surface scenarios have less neighbours \cite{becker2007weakly}. To correct this, we can use the continuity equation to obtain $\rho(\textbf{r}_i)$.

To derive the time derivative of density, assuming constant mass for each particle, we differentiate Eq.\ref{eq:particle_density} \textit{w.r.t} time:

\begin{equation} \label{eq:density_derivative1}
     \frac{d\rho_i}{dt} = \frac{d}{dt} \left( \sum_{j=1}^M m_j K_{ij} \right) 
     = \sum_{j=1}^M m_j \frac{d K_{ij}}{dt} 
\end{equation}

\noindent As the kernel function relies on the positions $r_i$ and $r_j$, the time derivative of the kernel function can be derived using the chain rule:

\[ 
    \frac{d K_{ij}}{dt} = \frac{d K_{ij}}{d \mathbf{r}_i} \cdot \frac{d\mathbf{r}_i}{dt} + \frac{d K_{ij}}{d \mathbf{r}_j} \cdot \frac{d\mathbf{r}_j}{dt}
\]

\noindent The velocities of particles \( i \) and \( j \) are \( \mathbf{v}_i = \frac{d\mathbf{r}_i}{dt} \) and \( \mathbf{v}_j = \frac{d\mathbf{r}_j}{dt} \). Using the fact that \( \nabla K_{ji} = -\nabla K_{ij} \), we have:

\[ \frac{d K_{ij}}{dt} = \nabla K_{ij}^T \cdot \mathbf{v}_i + \nabla K_{ji}^T \cdot \mathbf{v}_j = \nabla K_{ij}^T \cdot (\mathbf{v}_i - \mathbf{v}_j) \]

\noindent where we have denoted the spatial derivatives $\nabla K_{ij}=\nabla_{\textbf{r}} K(|\textbf{r}-\textbf{r}_j|,h) |_{\textbf{r}=\textbf{r}_i}$ and $\nabla K_{ji}=\nabla_{\textbf{r}} K(|\textbf{r}-\textbf{r}_i|,h) |_{\textbf{r}=\textbf{r}_j}$.

Therefore, Eq.\ref{eq:density_derivative1} translates to: 

\begin{equation} \label{eq:density_derivative2}
    \frac{d\rho_i}{dt} = \sum_{j=1}^M m_j (\mathbf{v}_i - \mathbf{v}_j)^T \cdot \nabla K_{ij}
\end{equation}

Eq.\ref{eq:density_derivative2} states that, each particle is initialized with a density $\rho_0$, and density changes are only due to relative motion of particles. It guarantees a correct density at free surfaces as well \cite{becker2007weakly}. One can choose to use Eq.\ref{eq:particle_density}, which is more stable but less accurate for surface particles, or solving Eq.\ref{eq:density_derivative2} to obtain the density.

In the following, we introduce two SPH formulations of the Navier–Stokes equations: weakly compressible SPH (WCSPH) and incompressible SPH (ISPH).

\paragraph{Weakly compressible SPH (WCSPH)}

The Navier-Stokes equations can be written in Lagrangian form as \cite{Lind2020}:

\begin{equation} \label{eq:NS_equation_WCSPH}
    \begin{aligned}
        \frac{d\rho}{dt} &= -\rho \nabla \cdot \mathbf{v} \quad \text{(Continuity equation)} \\
        \frac{d\mathbf{v}}{dt} &= -\frac{1}{\rho} \nabla P + \frac{1}{\rho}\mathbf{f}^{vis} + \frac{1}{\rho}\mathbf{f}^{ext} \quad \text{(Momentum equation)} \\
    \end{aligned}
\end{equation}

\noindent where $\mathbf{v}$ is the fluid velocity vector, $\rho$ is fluid density, $t$ is time, $P$ is fluid pressure. The first term on RHS of the momentum equation is the pressure gradient force, second term is the viscosity-induced resistance force (e.g. $\mathbf{f}^{vis}=\mu \nabla^2 \mathbf{v}$), and third term is all external (body) forces (e.g. gravity).

For example, if we ignore the viscous force (the momentum equation reduces to the Euler equation for inviscid
flow) and consider $\mathbf{f}^{ext}=\rho \mathbf{g}$, Eq.\ref{eq:NS_equation_WCSPH} becomes \cite{becker2007weakly}:

\begin{equation} \label{eq:NS_equation_WCSPH_gravity} \tag{\ref{eq:NS_equation_WCSPH}b}
    \begin{aligned}
        \frac{d\rho}{dt} &= -\rho \nabla \cdot \mathbf{v} \\
        \frac{d\mathbf{v}}{dt} &= -\frac{1}{\rho} \nabla P + \mathbf{g} \\
    \end{aligned}
\end{equation}

Particles conveniently carry properties such as mass, density and velocity; the pressure $P$, however, needs to be calculated. It can be derived from the EoS equation (Eq.\ref{eq:EoS_general}). In WCSPH, a stiff EoS, which relates fluid pressure to changes in density, is often used \cite{Lind2020}:

\begin{equation} \label{eq:EoS_pressure_density}
    P = \frac{c_0^2 \rho_0}{\gamma} \left[( \frac{\rho}{\rho_0} )^\gamma - 1 \right] \quad \text{(Equation of State)} \\
\end{equation}

\noindent where $\gamma$ is the fluid polytropic index, $\rho_0$ is the reference density, and $c_0$ is the reference numerical speed of sound in the fluid \cite{Lind2020,becker2007weakly}. Eq.\ref{eq:EoS_pressure_density} is derived based on \textit{the Tait equation} \footnote{The Tait equation can be used to efficiently enforce volume preservation as well \cite{monaghan1994simulating}.} \cite{monaghan1994simulating}.

To calculate the pressure $P_i$ at particle $i$, we apply SPH smoothing (Eq.\ref{eq:integral_discrete_approx}):

\begin{equation} \label{eq:pressure_SPH1}
    P(\textbf{r}_i) = \sum_{j=1}^M \frac{m_j}{\rho_j} P_j K(|\textbf{r}_i - \textbf{r}_j|,h)
\end{equation}

\noindent Eq.\ref{eq:pressure_SPH1} however is not symmetric; ideally, as per Newton's third law, we would like every action should have an equal and opposite reaction \cite{gibiansky2019computational}. This can be achieved by revising Eq.\ref{eq:pressure_SPH1} to be:

\begin{equation} \label{eq:pressure_SPH2} \tag{\ref{eq:pressure_SPH1}b}
    P(\textbf{r}_i) = \sum_{j=1}^M \frac{m_j}{\rho_j} \frac{P_i+P_j}{2} K(|\textbf{r}_i - \textbf{r}_j|,h)
\end{equation}

After obtaining $P$, we can apply the spatial gradient principle Eq.\ref{eq:diff_SPH} to obtain $\nabla P$:

\begin{equation} \label{eq:grad_pressure_SPH1}
    \nabla P(\textbf{r}_i) = \sum_{j=1}^M \frac{m_j}{\rho_j} P_j \nabla_{\textbf{r}} K(|\textbf{r}-\textbf{r}_j|,h) |_{\textbf{r}=\textbf{r}_i}
\end{equation}

\noindent or

\begin{equation} \label{eq:grad_pressure_SPH2} \tag{\ref{eq:grad_pressure_SPH1}b}
    \nabla P(\textbf{r}_i) = \sum_{j=1}^M \frac{m_j}{\rho_j} \frac{P_i+P_j}{2} \nabla_{\textbf{r}} K(|\textbf{r}-\textbf{r}_j|,h) |_{\textbf{r}=\textbf{r}_i}
\end{equation}

The discrete Lagrangian WCSPH Navier-Stokes equations can then be written as \footnote{\textit{NB}: Eq.\ref{eq:grad_pressure_SPH2} is read from \cite{gibiansky2019computational} while Eq.\ref{eq:discrete_NS_equation_WCSPH} is widely used in many literature such as \cite{Monaghan2005,mocz2011smoothed,Lind2020} - we need, however, more smoothing transition here to bridge both.} \cite{Monaghan2005}:

\begin{equation} \label{eq:discrete_NS_equation_WCSPH}
\begin{aligned}
    & \frac{d\rho_i}{dt} = \sum_{j=1}^M m_j (\mathbf{v}_i - \mathbf{v}_j)^T \cdot \nabla K_{ij} + \delta_i \quad \text{(Continuity equation)} \\
    & \frac{d\mathbf{v}_i}{dt} = -\sum_{j=1}^M m_j \left( \frac{P_i}{\rho_i^2} + \frac{P_j}{\rho_j^2} \right) \nabla K_{ij} + \frac{1}{\rho}\mathbf{f}_i^{vis} + \frac{1}{\rho}\mathbf{f}^{ext} \quad \text{(Momentum equation)}
    \\
\end{aligned}
\end{equation}

\noindent where $\mathbf{f}_i^{vis}$ denotes the viscous force. Different formulations of $\mathbf{f}_i^{vis}$ are possible, see Eq.\ref{eq:SPH_viscosity_force1} and Eq.\ref{eq:linear_viscosity_force}.

The final term in the density equation in Eq.\ref{eq:discrete_NS_equation_WCSPH}, $\delta_i$, is an artificial density diffusion term used to limit spurious density fluctuations (this SPH method is called $\delta$-SPH). One of the density diffusion schemes is suggested by \cite{Antuono2010}:

\begin{equation} \label{eq:artificial_density_diffusion}
\delta_i = a_d h c_0 \sum_{j=1}^{M} \frac{m_j}{\rho_j} \psi_{ij}^T \cdot \nabla K_{ij}
\end{equation}

\noindent where

\begin{equation*}
\psi_{ij} = 2 \left( \frac{\rho_j}{\rho_i} - 1 \right) \frac{\mathbf{r}_i - \mathbf{r}_j}{|\mathbf{r}_i - \mathbf{r}_j|^2 + 0.1 h^2}
\end{equation*}

\noindent where $a_d$ is a diffusion parameter,typically set to 0.1.

To discretize Eq.\ref{eq:discrete_NS_equation_WCSPH}, one can design a leapfrog integration scheme, which conserves energy, to update the position, velocity and density:

\begin{equation} \label{eq:leapfrog_integrator}
\begin{aligned}
    & \mathbf{v}_i^{(t+1)/2} = \mathbf{v}_i^t + \mathbf{a}_i^t \frac{\Delta t}{2}, \\
    & \mathbf{r}_i^{t+1} = \mathbf{r}_i^t + \mathbf{v}_i^{(t+1)/2} \Delta t, \\
    & \mathbf{v}_i^{t+1} = \mathbf{v}_i^{(t+1)/2} + \mathbf{a}_i^{t+1} \frac{\Delta t}{2}, \\
    & \rho_i^{t+1} = \rho_i^t + [\sum_{j=1}^M m_j (\mathbf{v}_i^{t+1} - \mathbf{v}_j^{t+1})^T \cdot \nabla K_{ij} + \delta_i] \Delta t \\
\end{aligned}
\end{equation}

\noindent where the superscript $t$ denotes the time step, $\mathbf{a}_i=d \mathbf{v}_i / dt$ is the acceleration which can be calculated via $\mathbf{a}_i=\mathbf{F}_i/m_i$ where $\mathbf{F}_i$ is the overall forces on the RHS of the momentum equation in Eq.\ref{eq:discrete_NS_equation_WCSPH}.

Note that, when applying the continuity equation in Eq.\ref{eq:discrete_NS_equation_WCSPH}, negative density values can occur due to the numerical integration process, the artificial density diffusion term and large step size $\Delta t$.

The WCSPH representation is summarised as follows:

\begin{mdframed}[frametitle={WCSPH equations}] \label{proc:WCSPH}
    \textbf{Navier-Stokes Equations (Lagrangian form):}
    \begin{equation*}
        \begin{aligned}
            & \frac{d\rho}{dt} = -\rho \nabla \cdot \mathbf{v} \\
            & \frac{d\mathbf{v}}{dt} = -\frac{1}{\rho} \nabla P + \frac{1}{\rho}\mathbf{f}_i^{vis} + \frac{1}{\rho}\mathbf{f}^{ext}, \\                        
        \end{aligned}
    \end{equation*}
    
    \textbf{Equation of State (EoS):}
    \begin{equation*}
        P = \frac{c_0^2 \rho_0}{\gamma} \left[ \left( \frac{\rho}{\rho_0} \right)^\gamma - 1 \right]
    \end{equation*}
    
    \textbf{Discrete SPH equations:}
    \begin{equation*}
        \begin{aligned}
            \frac{d\rho_i}{dt} &= \sum_{j=1}^M m_j (\mathbf{v}_i - \mathbf{v}_j)^T \cdot \nabla K_{ij} + \delta_i \\
            \frac{d\mathbf{v}_i}{dt} &= -\sum_{j=1}^M m_j \left( \frac{P_i}{\rho_i^2} + \frac{P_j}{\rho_j^2} \right) \nabla K_{ij} + \frac{1}{\rho}\mathbf{f}_i^{vis} + \frac{1}{\rho}\mathbf{f}^{ext}, \\
        \end{aligned}
    \end{equation*}
\end{mdframed}

\paragraph{Incompressible SPH (ISPH)} 

Incompressibility implies no volumetric variations, therefore density and mass are constant. The Navier-Stokes equations in Lagrangian form for an incompressible fluid become:

\begin{equation} \label{eq:NS_equation_ISPH}
\begin{aligned}
    & \nabla \cdot \mathbf{v} = 0 \quad \text{(Continuity equation)}
    \\
    & \frac{d\mathbf{v}}{dt} = -\frac{1}{\rho} \nabla P + \frac{1}{\rho}\mathbf{f}_i^{vis} + \frac{1}{\rho}\mathbf{f}^{ext} \quad \text{(Momentum equation)}
\end{aligned}
\end{equation}

\noindent where $\mathbf{v}$ again is the velocity, $\rho$ the density, $t$ is time, $P$ the pressure,  $\mathbf{f}^{vis}$ the viscous force, $\mathbf{f}^{ext}$ all external forces (e.g. gravity or other body forces). 

Again, if we consider $\mathbf{f}_i^{vis}=\mu \nabla^2 \mathbf{v}$ and $\mathbf{f}^{ext}=\rho\mathbf{g}$, Eq.\ref{eq:NS_equation_ISPH} become \cite{Lind2020}:

\begin{equation} \label{eq:NS_equation_ISPH_gravity} \tag{\ref{eq:NS_equation_ISPH}b}
\begin{aligned}
    & \nabla \cdot \mathbf{v} = 0
    \\
    & \frac{d\mathbf{v}}{dt} = -\frac{1}{\rho} \nabla P + \nu \nabla^2 \mathbf{v} + \mathbf{g}
\end{aligned}
\end{equation}

\noindent where $\nu=\mu/\rho$ is again the viscosity.  

As density remains constant and no longer needs to be solved, updating of the position and velocity can be proceeded as follows \cite{Lind2020}: at step $t$, for each particle $i$, we first calculate an intermediate position $\mathbf{r}_i^*$ (although not used) with velocity $\mathbf{v}_i^t$:
\[
\mathbf{r}_i^* = \mathbf{r}_i^t + \mathbf{v}_i^t \Delta t
\]

\noindent and an intermediate velocity $\mathbf{v}_i^*$ considering only the viscous force at time $t$ (e.g. $\frac{1}{\rho}\mathbf{f}_i^{vis,t} = \nu \nabla^2 \mathbf{v}_i^t$):
\[
\mathbf{v}_i^* = \mathbf{v}_i^t + \frac{1}{\rho} \mathbf{f}^{vis,t} \Delta t
\]

The pressure $P^{t+1}$ at time $t+1$ is obtained by the implicit solution of the Poisson equation \footnote{Discretization of the LHS of the above Poisson equation can be performed with either the Morris or Schwaiger operator.}:
\[
\nabla \cdot \left( \frac{1}{\rho} \nabla P^{t+1} \right) = \frac{1}{\Delta t} \nabla \cdot \mathbf{v}_i^*
\]

The velocity $\mathbf{v}_i^{t+1}$ at time $t+1$ is then obtained by considering the body force \footnote{Body forces, e.g. $\mathbf{f}^{ext}=\rho \textbf{g}$, are generally time invariant.} $\mathbf{f}^{ext}$ (e.g. $\mathbf{f}^{ext}=\rho \textbf{g}$):
\[
\mathbf{v}_i^{t+1} = \mathbf{v}_i^* + \left( -\frac{1}{\rho} \nabla P^{t+1} + \frac{1}{\rho}\mathbf{f}^{ext} \right) \Delta t,
\]

Finally particle positions $\mathbf{r}_i^{t+1}$ at time $t+1$ are calculated by central difference:
\[
\mathbf{r}_i^{t+1} = \mathbf{r}_i^t + \left( \frac{\mathbf{v}_i^t + \mathbf{v}_i^{t+1}}{2} \right) \Delta t.
\]

Note that, particle convergence can be a problem with ISPH, and regularization or shifting is essential to produce accurate converged solution \cite{Lind2020}. ISPH is summarised as below:

\begin{mdframed}[frametitle={ISPH equations}] \label{proc:ISPH}
    \textbf{Navier-Stokes equations (Lagrangian form):}
    \begin{equation*}
        \begin{aligned}
            & \nabla \cdot \mathbf{v} = 0, \\
            & \frac{d\mathbf{v}}{dt} = -\frac{1}{\rho} \nabla P + \frac{1}{\rho}\mathbf{f}_i^{vis} + \frac{1}{\rho}\mathbf{f}^{ext}
        \end{aligned}
    \end{equation*}
    
    \textbf{(1) Compute intermediate position $\mathbf{r}_i^*$ and velocity $\mathbf{v}_i^*$:}
    \begin{equation*}
        \begin{aligned}
            \mathbf{r}_i^* &= \mathbf{r}_i^t + \mathbf{v}_i^t \Delta t, \\
            \mathbf{v}_i^* &= \mathbf{v}_i^t + \frac{1}{\rho}\mathbf{f}_i^{vis,t} \Delta t
        \end{aligned}
    \end{equation*}
    
    \textbf{(2) Compute pressure $P^{t+1}$:}
    \begin{equation*}
        \nabla \cdot \left( \frac{1}{\rho} \nabla P^{t+1} \right) = \frac{1}{\Delta t} \nabla \cdot \mathbf{v}_i^*
    \end{equation*}
    
    \textbf{(3) Compute velocity $\mathbf{v}_i^{t+1}$ and position $\mathbf{r}_i^{t+1}$:}
    \begin{equation*}
        \begin{aligned}
            \mathbf{v}_i^{t+1} &= \mathbf{v}_i^* + \left( -\frac{1}{\rho} \nabla P^{t+1} + \frac{1}{\rho}\mathbf{f}^{ext} \right) \Delta t, \\
            \mathbf{r}_i^{t+1} &= \mathbf{r}_i^t + \frac{1}{2} (\mathbf{v}_i^t + \mathbf{v}_i^{t+1}) \Delta t
        \end{aligned}
    \end{equation*}
\end{mdframed}

\section{The smoothing kernel} \label{app:smoothing_kernel}

The quality, speed and accuracy of SPH simulation relies on the choice of the smoothing kernel $K(r,h)$ where \(r = |\mathbf{r}_i - \mathbf{r}_j|\) is the distance between two particles. For the SPH method, the smoothing kernel $K(r,h)$ is typically chosen to have a smooth and compact support. Here we present four commonly used kernels.

The gradient of the smoothing kernel $K(r,h)$, a directional vector, is defined as:

\[
\nabla K(r, h) = \frac{dK(r, h)}{dr} \cdot \frac{\mathbf{r}_i - \mathbf{r}_j}{r}
\]

\noindent where $\mathbf{r}_i$ and $\mathbf{r}_j$ are the positions of particle $i$ and $j$, respectively. The distance \(r = |\mathbf{r}_i - \mathbf{r}_j|\) is a scalar. \(h\) is the smoothing length. $(\mathbf{r}_i - \mathbf{r}_j)/r$ is the unit direction vector between particles particle $i$ and $j$. The smoothing kernel is stationary and  isotropic: it induces a ball centered at the point of concern with radius $h$, and the values $K(r, h)$ and $\nabla K(r, h)$ are homogeneous in all directions.

\paragraph{Empirical kernel} The general smoothing kernel is a sixth-degree polynomial

\begin{equation}
    K(r, h) = \frac{315}{64\pi h^9} (h^2 - r^2)^3
\end{equation}

\noindent and its gradient \textit{w.r.t} $r$:
\[
\frac{d K}{d r} = -\frac{945}{32 \pi h^9} \cdot r (h^2 - r^2)^2
\]

\noindent it avoids calculating the square root $r$ and is therefore faster \cite{gibiansky2019computational}.

\paragraph{Pressure kernel} Particles should repel more as they come closer, which inspires the following kernel for pressure calculation \cite{gibiansky2019computational}:

\begin{equation}
    K^{pre} (r, h) = \frac{15}{\pi h^6} (h - r)^3
\end{equation}

\noindent and its gradient \textit{w.r.t} $r$:
\[
\frac{d K^{\text{pre}}}{d r} = -\frac{45}{\pi h^6} (h - r)^2
\]

\paragraph{Viscosity kernel} Viscous force is a friction force and dissipates energy. The speed should not increase as particles come closer. This inspires the following kernel for viscosity calculation \cite{gibiansky2019computational}:

\begin{equation}
    K^{vis}(r, h) = -\frac{r^3}{2h^3} + \frac{r^2}{h^2} + \frac{h}{2r} - 1
\end{equation}

\noindent and its gradient \textit{w.r.t} $r$:
\[
\frac{d K^{\text{vis}}}{d r} = -\frac{3r^2}{2h^3} + \frac{2r}{h^2} - \frac{h}{2r^2}
\]

\paragraph{Cubic spline kernel} Another commonly used smoothing kernel is the cubic spline kernel (2D as an example):

\[
K(r, h) =
\begin{cases}
\frac{10}{7\pi h^2} \left(1 - \frac{3}{2} \left(\frac{r}{h}\right)^2 + \frac{3}{4} \left(\frac{r}{h}\right)^3 \right), & 0 \leq \frac{r}{h} \leq 1 \\
\frac{10}{7\pi h^2} \left(\frac{1}{4} \left(2 - \frac{r}{h}\right)^3 \right), & 1 < \frac{r}{h} \leq 2 \\
0, & \frac{r}{h} > 2
\end{cases}
\]

\noindent and its gradient:

\[
\frac{dK(r, h)}{dr} =
\begin{cases}
\frac{10}{7\pi h^3} \left(-3 \frac{r}{h} + \frac{9}{4} \left(\frac{r}{h}\right)^2 \right), & 0 \leq \frac{r}{h} \leq 1 \\
-\frac{10}{7\pi h^3} \left(\frac{3}{4} \left(2 - \frac{r}{h}\right)^2 \right), & 1 < \frac{r}{h} \leq 2 \\
0, & \frac{r}{h} > 2
\end{cases}
\]

\begin{figure}[h]
    \centering
    \begin{subfigure}{0.35\textwidth}
        \centering
        \includegraphics[width=\linewidth]{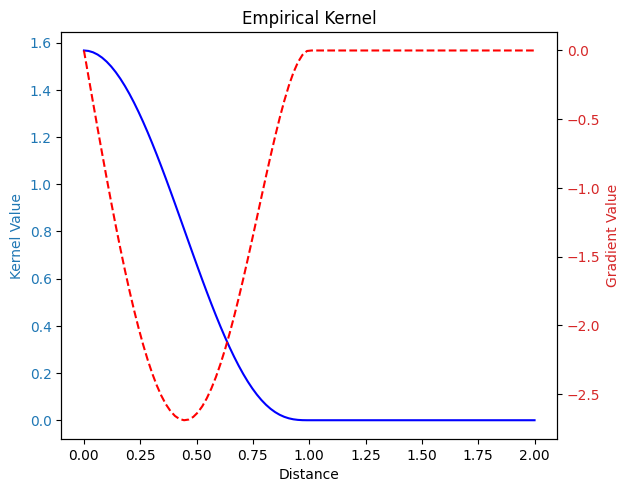}
        \caption{Empirical kernel}
        \label{fig:sub1}
    \end{subfigure}
    \begin{subfigure}{0.35\textwidth}
        \centering
        \includegraphics[width=\linewidth]{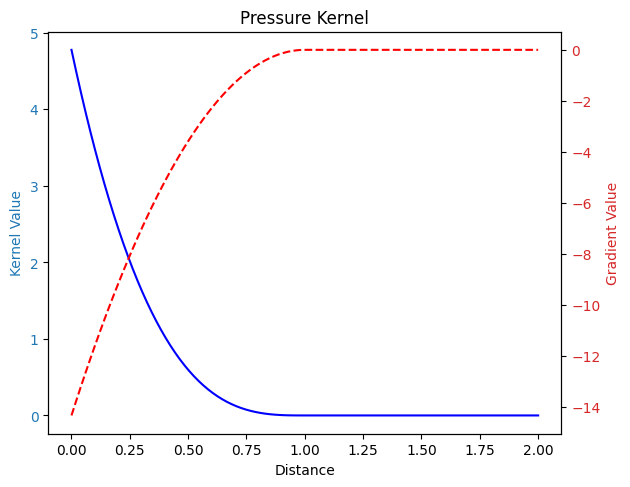}
        \caption{Pressure kernel}
        \label{fig:sub2}
    \end{subfigure}
    \\
    \begin{subfigure}{0.35\textwidth}
        \centering
        \includegraphics[width=\linewidth]{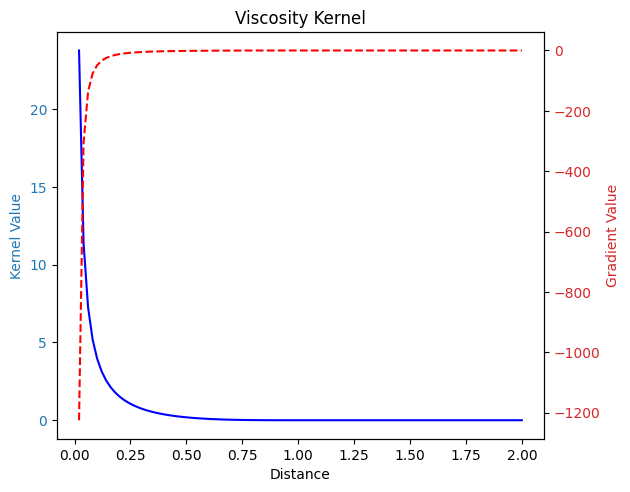}
        \caption{Viscosity kernel}
        \label{fig:sub3}
    \end{subfigure}
    \begin{subfigure}{0.35\textwidth}
        \centering
        \includegraphics[width=\linewidth]{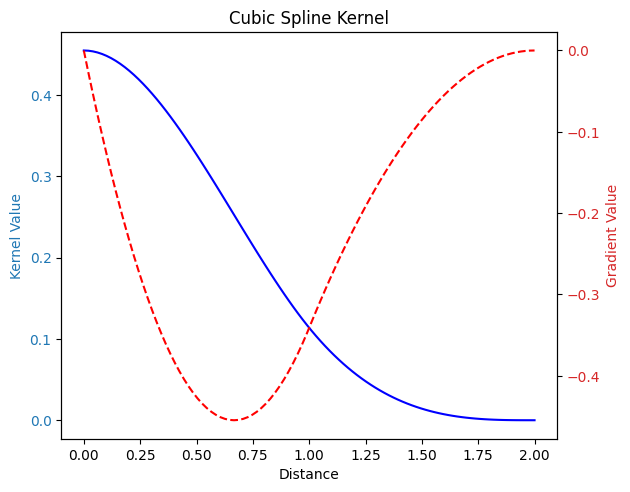}
        \caption{Cubic spline kernel}
        \label{fig:sub4}
    \end{subfigure}
    \caption{Smoothing kernels and gradients.}
    \label{fig:smoothing_kernels}
\end{figure}

The four kernels, along with their gradients, are shown in Fig.\ref{fig:smoothing_kernels}.

\section{SPH-based sampling (external force field)} \label{app:SPH_sampling_external_force_field}
The algorithm for the SPH-based sampling using an external force field is presented.

\begin{algorithm}[H]
\fontsize{8}{8}
\caption{SPH-based sampling (target score as external force field)}
\label{algo:SPH_sampling_external_force_field}
\begin{itemize}
\item{\textbf{Inputs}: a queryable target density $p(\textbf{r})$ with magnitude amplification constant $\alpha$, number of particles $M$, particle (density) dimension $d$, initial proposal distribution $p^0(\textbf{r})$, total number of iterations $T$, step size $\Delta t$, smoothing kernel $K(r,h)$ with length-scale $h$. Particle mass $m_i, i=1,2,...,M$. Reference density $\rho_0$, reference speed of sound in the fluid $c_0$, fluid polytropic index $\gamma$, dynamic viscosity $\mu$. Density diffusion parameter $a_d$.} 
\item{\textbf{Outputs}: Particles located at positions $\{\mathbf{r}_i \in \mathbb{R}^d \}_{i=1}^{M}$ whose empirical distribution approximates the target density $p(\textbf{r})$.}
\end{itemize}
\vskip 0.06in
1. \textbf{\textit{Initialise}} particles [$\mathcal{O}(M)$].
    \begin{addmargin}[1em]{0em}%
    Draw $M$ $m$-dimensional particles $\textbf{r}^0=\{\mathbf{r}_j^0\}_{j=1}^{M}$ from the initial proposal distribution $p^0(\textbf{r})$. Initialise $(\rho_{i}^0,\mathbf{r}_{i}^0,\mathbf{v}_{i}^0)$ for i=1,2,...,M. \\ 
    \end{addmargin}

2. \textbf{\textit{Update}} particle positions. \\
For each iteration $t=1,2,...,T$, repeat until converge:
    \\
    \begin{addmargin}[1em]{0em}%
    (1) Compute the density $\rho_i^t$ for each particle $i$=1,2,...,M (Eq.\ref{eq:simplified_density_cal} or \ref{eq:cc_leapfrog_integrator}) [$\mathcal{O}(M^2)$]:
    \[
    \rho_i^t = \sum_{j=1}^M m_j K(|\mathbf{r}_i^t - \mathbf{r}_j^t|, h), 
    \text{ or }
    \rho_i^{t} = \rho_i^{t-1} + [\sum_{j=1}^M m_j (\mathbf{v}_i^{t} - \mathbf{v}_j^{t})^T \cdot \nabla K_{ij} + \delta_i^{t}] \Delta t
    \]
    where $\delta_i^{t}=a_d h c_0 \sum_{j=1}^{M} m_j \frac{{\psi_{ij}^t}^T}{\rho_j^{t}} \cdot \nabla K_{ij}$ and $\psi_{ij}^{t} = 2 \left( \frac{\rho_j^{t}}{\rho_i^{t}} - 1 \right) \frac{\mathbf{r}_i^{t} - \mathbf{r}_j^{t}}{|\mathbf{r}_i^{t} - \mathbf{r}_j^{t}|^2 + 0.1 h^2}$. \\ 
    \\
    (2) Compute the (internal) pressure $P_i^t$, and external forces $\mathbf{f}_i^{ext}$ by querying $\nabla_{\mathbf{r}} p(\mathbf{r}_i^t)$, for each particle $i$ [$\mathcal{O}(M)$]:
    \[
    P_i^t = \frac{c_0^2 \rho_0}{\gamma} [( \frac{\rho_i^t}{\rho_0} )^\gamma - 1]
    , \quad 
    \mathbf{f}_{i}^{ext,t} = \alpha \nabla_{\mathbf{r}} \log p(\mathbf{r}_i^t)
    \]
    \\
    (3) Compute the overall force $\mathbf{F}_i^t$ acting on each particle $i$ (Eq.\ref{eq:external_force_filed_momentum_equation}) [$\mathcal{O}(M^2)$]:  
    \[ 
    \mathbf{F}_i^t = -\sum_{j=1}^M m_j \left( \frac{P_i^t}{\rho_i^{t2}} + \frac{P_j^t}{\rho_j^{t2}} \right) \nabla K_{ij} + \mathbf{f}_i^{vis,t} + \mathbf{f}_{i}^{ext,t}
    \]
    where $\mathbf{f}_i^{vis,t}$ is calculated using Eq.\ref{eq:SPH_viscosity_force1} or Eq.\ref{eq:linear_viscosity_force}. \\ 
    \\
    (4) Update $(\mathbf{r}_{i}^t,\mathbf{v}_{i}^t)$ (\ref{eq:cc_leapfrog_integrator} simplified) [$\mathcal{O}(M)$]: \\
    \[
     \mathbf{v}_i^{t+1} = \mathbf{v}_i^t + \frac{\mathbf{F}_i^t}{m_i} \Delta t, \hspace{0.25cm} \mathbf{r}_i^{t+1} = \mathbf{r}_i^t + \mathbf{v}_i^{t+1} \Delta t
    \]    
    \end{addmargin}

3. \textbf{\textit{Return}} the final configuration of particles $\{\mathbf{r}_j^{T}\}_{j=1}^{M}$ and their histogram and/or \textit{KDE} estimate for each dimension. \\
\end{algorithm}

\paragraph{Practical implementation} Based on the above algorithm, in order to optimise efficiency, at each iteration $t$, following procedure can be implemented:

\begin{itemize}
    \item \textit{Step 1}: compute and cache distance matrix $R^t_{M \times M}$ (symmetric, with entries $r^t_{ij}=|\textbf{r}_i^t - \textbf{r}_j^t|$), kernel value matrix $K^t_{M \times M}$ (symmetric, with entries $k^t_{ij}=K(|\textbf{r}_i^t- \textbf{r}_j^t|,h)$), kernel gradient tensor $\nabla K^t_{M \times M \times d}$ (asymmetric, with entries $\nabla k^t_{ij}= \nabla K(|\textbf{r}_i^t - \textbf{r}_j^t|,h)$).
    
    \item \textit{Step 2}: compute particle densities $\{\rho (\textbf{r}_i^t)=\sum_{j=1}^M m_j K(|\mathbf{r}_i^t - \mathbf{r}_j^t|, h)\}_{i=1}^M$ for each particle, and query target value $\{\nabla_{\textbf{r}} \log p(\textbf{r}_i^t) \}_{i=1}^M$ at each particle position.

    \item \textit{Step 3}: retrieve densities and compute internal pressures $\{ P_i^t = \frac{c_0^2 \rho_0}{\gamma} [( \frac{\rho_i^t}{\rho_0} )^\gamma - 1] \}_{i=1}^M$.

    \item \textit{Step 4}: retrieve densities, pressures and distances, compute overall force acting on each particle $\{ \textbf{F}_i^t = \textbf{f}_i^{pre,t} +\textbf{f}_i^{vis,t} +\textbf{f}_i^{ext,t} \}_{i=1}^M$, where $\textbf{f}_i^{pre,t}=-\sum_{j=1}^M m_j \left( \frac{P_i^t}{\rho_i^{t2}} + \frac{P_j^t}{\rho_j^{t2}} \right) \nabla K_{ij}$, $\mathbf{f}_i^{vis,t} = \sum_{j=1}^M \frac{\mu}{\rho_j^t} \frac{m_j}{\rho_j^t} \mathbf{v}_j^t \nabla^2 K(|\textbf{r}_i^t - \textbf{r}_j^t|,h)$ or $\textbf{f}_i^{vis, t}=m_j \eta_{ij} \frac{(\mathbf{v}_i^t - \mathbf{v}_j^t)^T \cdot (\mathbf{r}_i^t - \mathbf{r}_j^t)}{|\mathbf{r}_i^t - \mathbf{r}_j^t|^2 + \epsilon \cdot h^2} \nabla K(|\mathbf{r}_i^t - \mathbf{r}_j^t|, h)$, $\textbf{f}_i^{ext,t}=\alpha \nabla_{\textbf{r}} \log p(\textbf{r}_i^t)$.

    \item \textit{Step 5}: retrieve forces, update velocities and positions $\{ \mathbf{v}_i^{t+1} = \mathbf{v}_i^t + \frac{\mathbf{F}_i^t}{m_i} \Delta t, \hspace{0.25cm} \mathbf{r}_i^{t+1} = \mathbf{r}_i^t + \mathbf{v}_i^{t+1} \Delta t \}_{i=1}^M$.
\end{itemize}

\end{document}